\def\cvprPaperID{2966}
\def\confName{CVPR}
\def\confYear{2023}

\def\paperTitle{Generalizing Dataset Distillation via Deep Generative Prior}

\def\authorBlock{
George Cazenavette\textsuperscript{1}\;\;\;\;\;Tongzhou Wang\textsuperscript{1}\;\;\;\;\;Antonio Torralba\textsuperscript{1}\;\;\;\;\;Alexei A. Efros\textsuperscript{2}\;\;\;\;\;Jun-Yan Zhu\textsuperscript{3}\\
\\
\textsuperscript{1}Massachusetts Institute of Technology\;\;\;\;\;\textsuperscript{2}UC Berkeley\;\;\;\;\;
\textsuperscript{3}Carnegie Mellon University
\vspace{0.2cm}
\\
\small{
\href{https://georgecazenavette.github.io/glad/}{\color{blue}georgecazenavette.github.io/glad}
}
\vspace{-0.3cm}
}

\newif\ifreview 
\newif\ifarxiv \newcommand{\arxiv}{\arxivtrue}
\newif\ifcamera 
\newif\ifrebuttal 

\arxiv

\pdfoutput=1
\documentclass[10pt,twocolumn,letterpaper]{article}

\PassOptionsToPackage{table,xcdraw}{xcolor}
\input{cvpr_header}

\begin{document}
\title{\paperTitle}
\author{\authorBlock}
\newcommand\coverwidth{0.093}

\twocolumn[{
\maketitle
\ssmall
\begin{center}
\begingroup
\setlength{\tabcolsep}{1pt}
\vspace*{-0.5cm}
\begin{tabular}{ccccccccccc}
\rotatebox[origin=c]{90}{\scriptsize{ImageNet-Birds}} &
    \includegraphics[align=c,width=\coverwidth\linewidth]{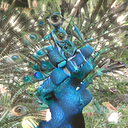} &
    \includegraphics[align=c,width=\coverwidth\linewidth]{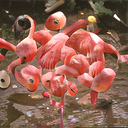} &
    \includegraphics[align=c,width=\coverwidth\linewidth]{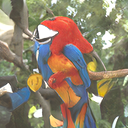} &
    \includegraphics[align=c,width=\coverwidth\linewidth]{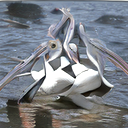} &
    \includegraphics[align=c,width=\coverwidth\linewidth]{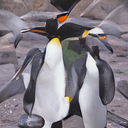} &
    \includegraphics[align=c,width=\coverwidth\linewidth]{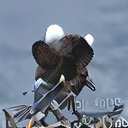} &
    \includegraphics[align=c,width=\coverwidth\linewidth]{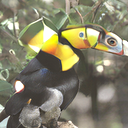} &
    \includegraphics[align=c,width=\coverwidth\linewidth]{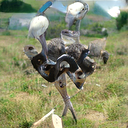} &
    \includegraphics[align=c,width=\coverwidth\linewidth]{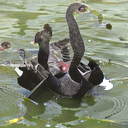} &
    \includegraphics[align=c,width=\coverwidth\linewidth]{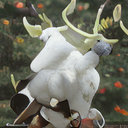} \\[7.8ex]
    & Peacock & Flamingo & Macaw & Pelican & Penguin & Eagle & Toucan & Ostrich & Black Swan & Cockatoo\\[-0.15cm]\\
    \rotatebox[origin=c]{90}{\scriptsize{ImageNet-Fruits}} &
    \includegraphics[align=c,width=\coverwidth\linewidth]{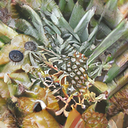} &
    \includegraphics[align=c,width=\coverwidth\linewidth]{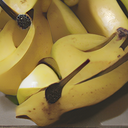} &
    \includegraphics[align=c,width=\coverwidth\linewidth]{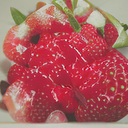} &
    \includegraphics[align=c,width=\coverwidth\linewidth]{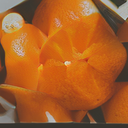} &
    \includegraphics[align=c,width=\coverwidth\linewidth]{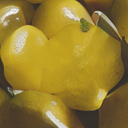} &
    \includegraphics[align=c,width=\coverwidth\linewidth]{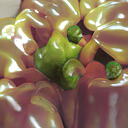} &
    \includegraphics[align=c,width=\coverwidth\linewidth]{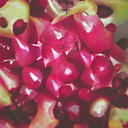} &
    \includegraphics[align=c,width=\coverwidth\linewidth]{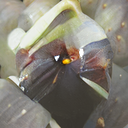} &
    \includegraphics[align=c,width=\coverwidth\linewidth]{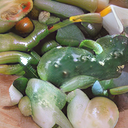} &
    \includegraphics[align=c,width=\coverwidth\linewidth]{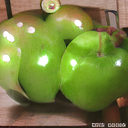}\\[7.8ex]
    &Pineapple & Banana & Strawberry & Orange & Lemon & Bell Pepper & Pomegranate & Fig & Cucumber & Granny Smith\\[-0.15cm]\\
    \rotatebox[origin=c]{90}{\scriptsize{ImageNet-Cats}} &
    \includegraphics[align=c,width=\coverwidth\linewidth]{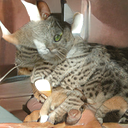} &
    \includegraphics[align=c,width=\coverwidth\linewidth]{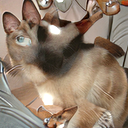} &
    \includegraphics[align=c,width=\coverwidth\linewidth]{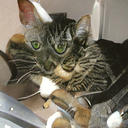} &
    \includegraphics[align=c,width=\coverwidth\linewidth]{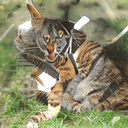} &
    \includegraphics[align=c,width=\coverwidth\linewidth]{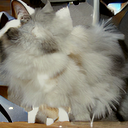} &
    \includegraphics[align=c,width=\coverwidth\linewidth]{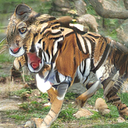} &
    \includegraphics[align=c,width=\coverwidth\linewidth]{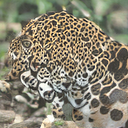} &
    \includegraphics[align=c,width=\coverwidth\linewidth]{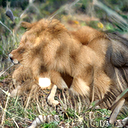} &
    \includegraphics[align=c,width=\coverwidth\linewidth]{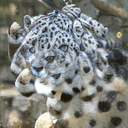} &
    \includegraphics[align=c,width=\coverwidth\linewidth]{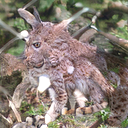} \\[7.8ex]
    & Egyptian & Siamese & Tabby & Bengal & Persian & Tiger & Jaguar & Lion & Snow Leopard & Lynx 
\end{tabular}
\endgroup\vspace{-0.25cm}
    \captionof{figure}{Visualizations of distilled images from three ImageNet subsets. Rather than distilling into pixel space, our \textit{Deep Generative Prior} constrains the images to be more coherent, leading to better cross-architectural generalization.}
    \label{fig:teaser}
\end{center}
}]

\begin{abstract}
Dataset Distillation aims to distill an entire dataset's knowledge into a few synthetic images. 
The idea is to synthesize a small number of synthetic data points that, when given to a learning algorithm as training data, result in a  model approximating one trained on the original data. 
Despite recent progress in the field, existing dataset distillation methods fail to generalize to new architectures and scale to high-resolution datasets. 
To overcome the above issues, we propose to use the learned prior from pre-trained deep generative models to synthesize the distilled data. 
To achieve this, we present a new optimization algorithm that distills a large number of images into a few intermediate feature vectors in the generative model's latent space. 
Our method augments existing techniques, significantly improving cross-architecture generalization in all settings.
\end{abstract}
\begin{figure*}[tp]
    \centering
    \includegraphics[width=0.8\linewidth, trim=0 0 0 0, clip]{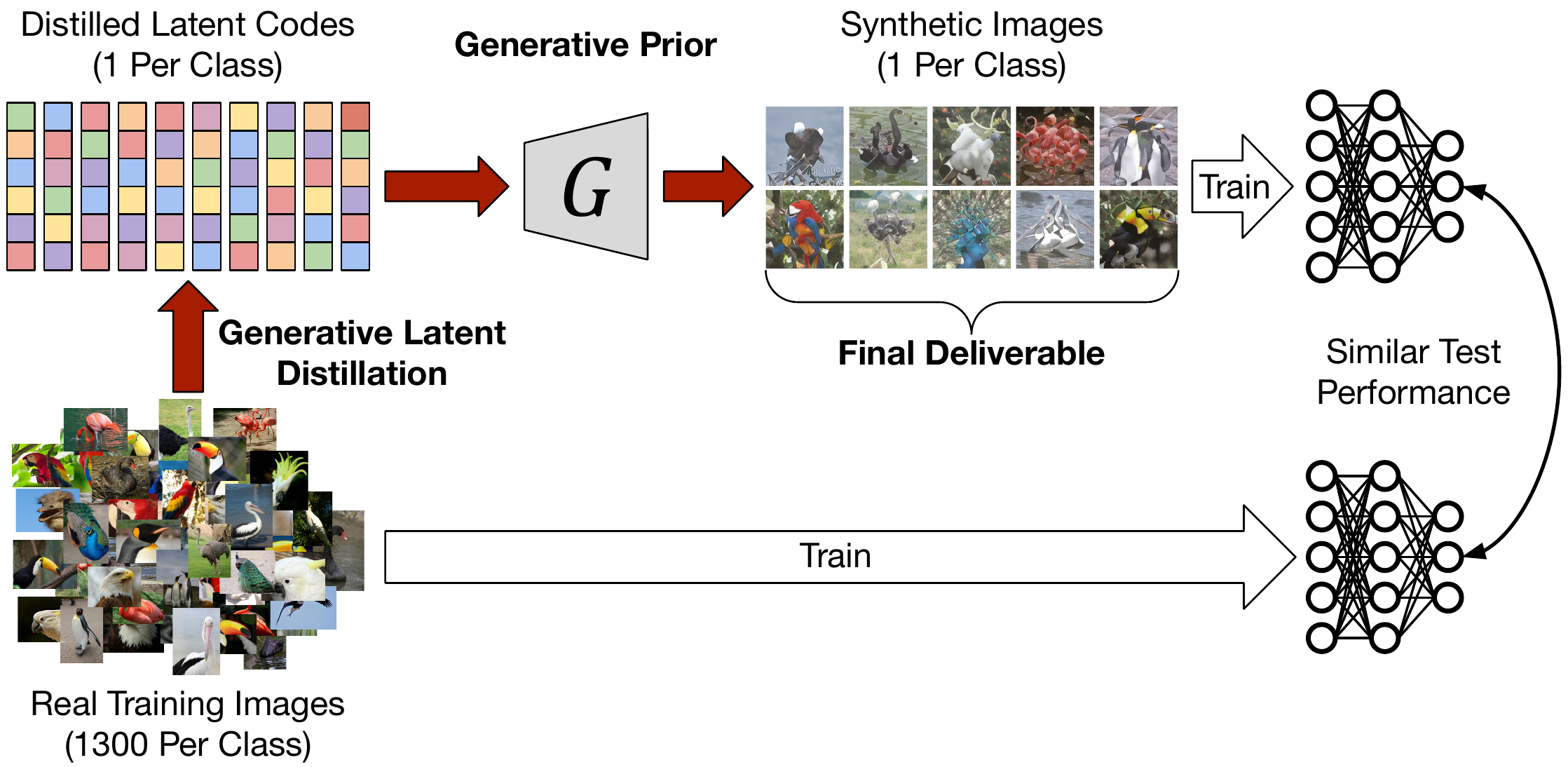}%
    \caption{Rather than directly distilling a dataset into synthetic pixels (like all previous methods), our new method \texttt{GLaD} instead distills into the latent space of a \textit{deep generative prior}. This enforces a tuneable amount of coherence in the output synthetic images, leading to far better generalization to new architectures.}
    \label{fig:method}
\end{figure*}
\section{Introduction}
\label{sec:intro}
\looseness=-1
Many recent advancements in machine learning come from combining large networks and big data. 
Such trained models have shown strong capabilities to perform a wide range of diverse tasks \cite{brown2020language,ramesh2022hierarchical,devlin2018bert}. 
Despite the great potential of such approaches, we, as a scientific community, are also curious about the underlying principles and limitations. 
Do networks have to be large to express the functions of interest? Do datasets have to be big? Can training on ``small data'' be equally successful? 

\looseness=-1
The seminal work on Knowledge Distillation \cite{hinton2015distilling} and recent discoveries such as Lottery Ticket Hypothesis \cite{frankle2018lottery} have revealed small models are often sufficient to approximate the same functions as large trained models. %
Dataset Distillation~\cite{dd} investigates the analogous yet orthogonal question on datasets: is there a small succinct dataset sufficient for training models? 
In other words, Dataset Distillation aims to distill a large dataset into a small (synthetic) one, such that training on the small dataset yields comparable performance (\reffig{method}). 
Since its introduction, Dataset Distillation has gained much attention in the research community, leading to many applications \citep{masarczyk2020reducing,such2020generative,dong2022privacy,cazenavette2022distillation}, and a growing series of methods that address the distillation problem: generating a discrete set of images effective for model training~\cite{dd,dc,dsa,dm,mtt,wang2022cafe,frepo}. 
When optimizing for a small, synthetic vision dataset, such methods typically optimize the \textit{raw} pixel values of the images.

\looseness=-1
Unfortunately, these methods face two major challenges, limiting both their scientific value and empirical applications. 
First, the distilled synthetic dataset is often optimized \wrt a specific network architecture, but struggles to generalize to other architectures. 
Second, while producing insightful distilled images on toy datasets, these methods generally fail to work well on higher-resolution datasets (\eg $\geq 128\times128$ resolution) and tend to distill visually noisy images with subpar performance. 

In this work, we argue that both issues are partially caused by parameterizing the synthetic dataset in pixel space. 
Directly optimizing pixels can be susceptible to learning high-frequency patterns that overfit the specific architecture used in training. 
To address this, we consider regularizing the distillation process to some prior that may help cross-architecture generalization. 
However, how and where to perform this regularization poses a delicate balance. 
For example, restricting our synthetic set to the real data manifold can significantly reduce the cross-architecture performance gap but is too strong a regularization to learn good distilled datasets. 
In the limit, it reduces to coreset selection~\cite{tsang2005core,borsos2020coresetscore,har2007smallercore,baykal2018smallcore}, which is known to not work as well~\cite{dd,dc,dsa,mtt}.

\looseness=-1
We propose Generative Latent Distillation (\texttt{GLaD}), which employs a \emph{deep generative prior} by parameterizing the synthetic dataset in the \emph{intermediate} feature space of generative models, such as Generative Adversarial Networks (GANs) \cite{goodfellow2014generative}. 
Our prior encourages the learned datasets to be more generalizable to new architectures but is also lax enough to \emph{not} prohibitively restrict the expressiveness of the distilled dataset. \texttt{GLaD} acts as an add-on module and can easily be applied to existing and future dataset distillation methods.

There is flexibility in choosing which generative model to use as our prior. 
Using a generator trained on the same target dataset as the distillation algorithm, our prior uses no additional data or information but consistently improves various distillation algorithms. 
However, to obtain the best distilled synthetic dataset, we may use more powerful generators trained on larger datasets. 
On the other extreme, we explore using randomly initialized generators and generators trained on out-of-distribution datasets. We show that they generate aesthetically pleasing synthetic images with distinct visual characteristics and achieve comparable distillation performance.  
In short, while different generator choices affect distillation results in intriguing ways, \texttt{GLaD} consistently improves performance across many datasets and multiple distillation algorithms. 

Within a deep generative model, there is a spectrum of different latent space choices corresponding to different layers in the model \citep{abdal2019image2stylegan,parmar2022sam,zhu2021barbershop}. Our analysis reveals a trade-off between realism (earlier layer latents) and flexibility (later layer latents), and highlights that using an intermediate latent space achieves a nice balance and consistent performance gain compared to the wildly used raw-pixel parametrization.

In \refsec{expr}, we perform extensive experiments on CIFAR-10 and ImageNet subsets at resolutions up to $512 \times 512$. We integrate \texttt{GLaD} with three distinct distillation algorithms (Gradient Matching \cite{dc}, Distribution Matching \cite{dm}, and Trajectory Matching \cite{mtt}), and consistently observe significant improvements in cross-architecture generalization. Our analysis of results from different configurations provides a better understanding of the effect of different generative models and latent spaces. 
Additionally, our method drastically reduces the high-frequency noise present in high-resolution datasets distilled into pixel space, leading to visually pleasing images that may have implications in artistic and design applications~\cite{cazenavette2022distillation}.

\texttt{GLaD} is a plug-and-play addition to existing and future distillation methods, allowing them to scale up to more realistic datasets and generalize better to different architectures. Given the goal of distilling large-scale datasets, we propose that leveraging generative models and differentiable parameterizations is a natural path forward. Our code and further visualizations be found on our project page \url{https://georgecazenavette.github.io/glad/}.

\section{Related Work}
\label{sec:related}
\looseness=-1
\textbf{Dataset Distillation} was introduced by Wang et al.~\citep{dd} as an investigation as to how training on very little data could update a model to certain desired behaviors. Several improvements have been made since then, including soft learned labels \citep{sucholutsky2021soft}, data augmentation \citep{dsa}, and trajectory/gradient matching \citep{mtt,dc}. Many applications have also been explored, such as neural architecture search \citep{ho2016generative} and continual learning \citep{dc,dsa}. Some works \citep{huang2021generative,ho2016generative} also employ generative models but train them from scratch to produce good synthetic training samples, in a manner different from our objective and formulation.

Several concurrent works tackle the dataset distillation via a re-parameterization of the distilled data as a set of bases \cite{deng2022remember,liu2022dataset} or a greater number of lower-resolution images \cite{kim2022dataset}, motivated by a memory compression standpoint. Our method differs in that we use generative model parameterizations to achieve better distillation performance. Concurrent work \cite{zhao2022synthesizing} also learns synthetic datasets in the latent space of a generative model by first inverting the \textit{entire} training set into this latent space before fine-tuning the latents on the distillation objective.

\myparagraph{GAN Inversion and Latent Spaces.} 
Our work applies generative priors by parameterizing distilled images into different latent spaces of GANs. This is related to the line of research of GAN inversion~\citep{zhu2016generative,brock2017neural}, whose goal is to project a real input image to GAN latent codes for image editing and data augmentation purposes~\cite{tewari2020pie,zhu2020domain,huh2020transforming,wei2021simpleinversion,chai2021using,chai2021ensembling,wang2022high}. 
These works have proposed various latent spaces (roughly similar to different activation spaces) \citep{abdal2019image2stylegan,abdal2020image2stylegan++,wu2020stylespace,zhu2021barbershop,harkonen2020ganspace}, wherein image editing or model fine-tuning can be efficiently performed \citep{richardson2021encoding,alaluf2021restyle,bau2020semantic,bau2020rewriting,roich2021pivotal,pan2021exploiting,parmar2022sam}.

The choice of specific latent space represents a trade-off between expressiveness and reconstruction quality, as shown by recent studies~\cite{tov2021designing,zhu2020improved}. In our experiments, we leverage a spectrum of such latent spaces designed for the StyleGAN-family of models~\citep{karras2019style,karras2020analyzing,stylegan2ada,sgxl} and discover that such a trade-off also exists in our case. The ideal distilled images are not realistic but can benefit from some level of generative prior. Our experiments perform an in-depth analysis on this aspect and show that an \textit{intermediate} generative prior (\ie starting from an intermediate layer) yields the best performance.

\begin{figure*}[]
    \centering
    \hspace*{-0.02\linewidth}\includegraphics[width=\linewidth, trim=20 0 20 0, clip]{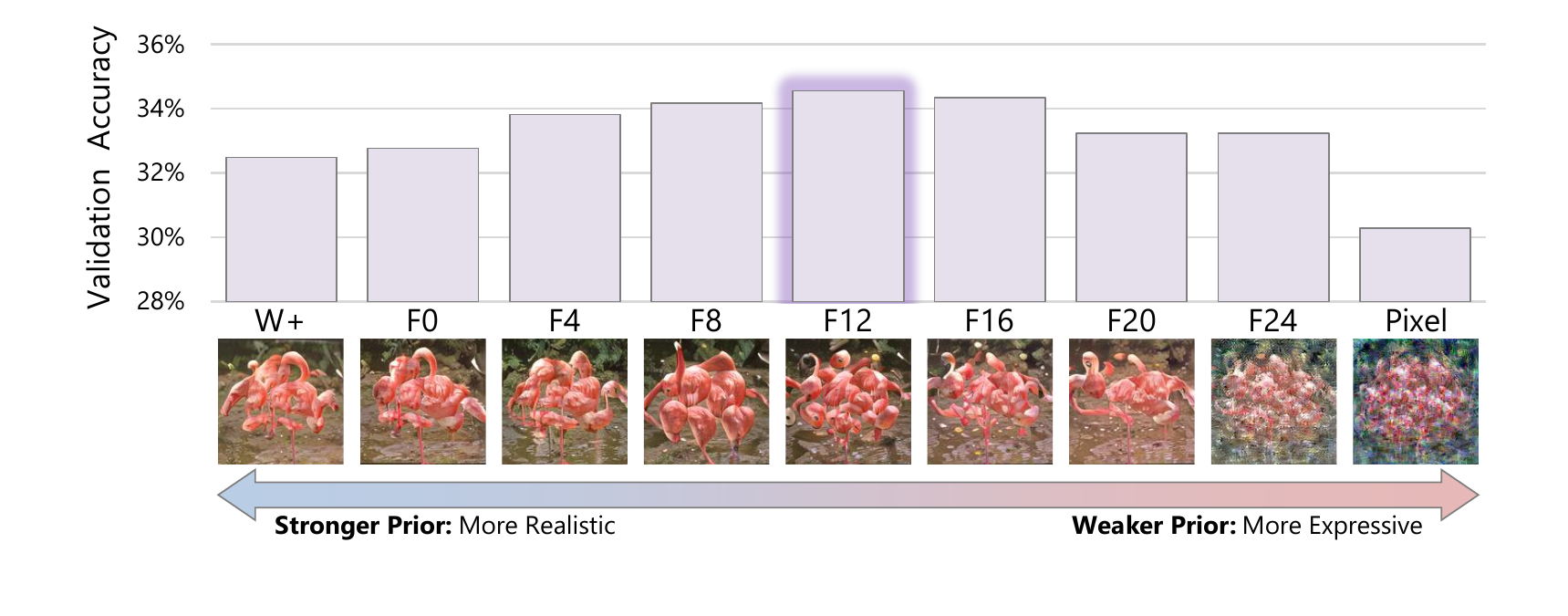}\vspace{-30pt}
    \caption{
    \textbf{Cross-Architecture performance of \mtt{} \cite{mtt} per distillation space averaged over ImageNet-[A,B,C,D,E].} Using spaces of earlier layers (\textbf{left side}) imposes stronger priors and yields more realistic images but is not as expressive as using spaces of later layers (\textbf{right side}). A proper amount of generative prior is helpful for cross-architecture generalization, while too strong a prior limits expressivity and thus hurts distillation performance. An intermediate space (\eg F12) balances this trade-off.}
    \lblfig{fig:layers}
\end{figure*}
\newcommand{\COMMENT}[2][.485\linewidth]{%
  \leavevmode\hfill\makebox[#1][l]{$\triangleright$~#2}}
\begin{algorithm}[t]
    \small
    \caption{Generative Latent Distillation}
    \label{alg:alg}
    \begin{algorithmic}[1]
        \Require {$\mbox{Alg}$: Distillation algorithm (\mtt{}, \dc{}, or \dm{}).}
        \Require {$\mathcal{T}$: Real training set.}
        \Require {$\mathcal{A}$: Differentiable augmentation function.}
        \Require {$G$: Pre-trained generator.}
        \Require {$P_z$: Distribution of latent initializations.}
        \State $\mathcal{Z}\sim P_z$\COMMENT{Initialize distilled latents}
        \For {\textbf{each} distillation step...}
         \label{step:outer}
            \State $\mathcal{S} = G(\mathcal{Z})$\COMMENT{Get images from latents}
            \State $\mathcal{L} =\mbox{Alg}(\mathcal{S}, \mathcal{T})$ \COMMENT{Compute distillation loss}
            \State $\mathcal{Z} \leftarrow \mbox{SGD}(\mathcal{Z}; \mathcal{L})$ \COMMENT{Update $\mathcal{Z}$ with respect to $\mathcal{L}$}
        \EndFor
        \Ensure {Distilled images $\mathcal{S}=G(\mathcal{Z})$}
    \end{algorithmic}
        \lblalg{main}
\end{algorithm}

\section{Generative Latent Distillation (\texttt{GLaD})}
\label{sec:method}
To date, all existing methods of dataset distillation \cite{dd,bohdal2020flexible,dc,dsa,nguyen2020dataset,nguyen2021dataset,wang2022cafe,mtt} have relied on a ``backbone'' architecture to formulate the distillation objective. 
Since optimizing distilled images in pixel space allows too much freedom to overfit to the backbone architecture, we propose introducing a \textit{deep generative prior} to the distillation process as a form of regularization by optimizing latent codes of a pre-trained generative model rather than raw pixels. We call our method \textit{Generative Latent Distillation} (\texttt{GLaD}). 

\subsection{Preliminaries on Dataset Distillation Methods}\lblsec{TM}
For completeness, we briefly review the three dataset distillation methods on which we conduct experiments: Gradient Matching (\dc)~\cite{dc}, Distribution Matching (\dm)~\cite{dm}, and Matching Training Trajectories (\mtt)~\cite{mtt}.
These three methods all seek to distill a small synthetic dataset $\mathcal{S}$ such that a model trained from scratch on $\mathcal{S}$ will have similar performance as a model trained on the full, real dataset $\mathcal{T}$.

\textbf{Dataset Condensation} (\dc) enforces that the gradients of the classification loss $\ell$ w.r.t. the synthetic images match those of the real images. 
For a network with parameters $\theta$ trained on the synthetic data for some number of iterations, the gradient matching loss is
\begin{equation}
    \mathcal{L}_{\dc} =1- \frac{\nabla_\theta\ell^\mathcal{S}(\theta) \cdot \nabla_\theta\ell^\mathcal{T}(\theta)}{\left\|\nabla_\theta\ell^\mathcal{S}(\theta)\right\|\left\|\nabla_\theta\ell^\mathcal{T}(\theta)\right\|}.
\end{equation}

\textbf{Distribution Matching} (\dm) takes a different approach and instead requires that a feature extractor yields similar output for real and synthetic images of a corresponding class. 
For a randomly initialized feature extractor $\psi$, the distribution matching loss is 
\begin{equation}
    \mathcal{L}_{\dm} = \sum_{c}\left\|\frac{1}{|\mathcal{T}_c|}\sum_{\mathbf{x}\in \mathcal{T}_c}\psi(\mathbf{x})-\frac{1}{|\mathcal{S}_c|}\sum_{\mathbf{s}\in \mathcal{S}_c}\psi(\mathbf{s})\right\|^2,
\end{equation}
where $\mathcal{T}_c, \mathcal{S}_c$ are the real and synthetic images for class $c$. 

\looseness=-1
Both the \dc{} and \dm{} methods are augmented with Differential Siamese Augmentation (\dsa)~\cite{dsa} such that the same differential augmentations~\cite{diffaug,stylegan2ada} are applied to real and synthetic images at each iteration. %

\textbf{Matching Training Trajectories} (\mtt) focuses on consistent training over a longer time horizon by encouraging the synthetic data to induce similar update trajectories in parameter space.
Prior to distillation, many \textit{expert trajectories} $\{\theta_t^*\}_0^T$ are obtained by training networks from scratch on the full real dataset and storing parameter snapshots at given intervals.
At each distillation step, a random expert trajectory and starting timestamp $\theta_t^*$ are sampled.
A student network is then initialized at the given expert timestamp $\hat{\theta}_t \coloneqq \theta_t^*$ and trained for $N$ iterations on the \textit{synthetic data}.
The trajectory-matching loss is then calculated as the normalized mean-squared error between the final parameters of the student network $\hat{\theta}_{t+N}$ and those of a future timestep ($M$ steps ahead) of the expert trajectory $\theta^*_{t+M}$:
\begin{equation}
  \mathcal{L}_{\mtt} = \frac{\|\hat{\theta}_{t+N} - \theta^*_{t+M}\|^2}{\|\theta^*_{t} - \theta^*_{t+M}\|^2}.  
\end{equation}
\looseness=-1
We make use of a recent method, dubbed \tesla{} \cite{tesla}, which solves the memory issue of \mtt{} by re-formulating the loss function to an equivalent form that only requires the storing of one set of network gradients at a time, thereby reducing the memory usage from linear to constant w.r.t. the number of inner loops $N$. However, we do not use the method's soft label assignment.
\subsection{\textbf{\texttt{GLaD}}: Adding a Deep Generative Prior}\lblsec{DGP}
Rather than naively optimizing the synthetic pixels directly (as in all previous methods \cite{dd,bohdal2020flexible,dc,dsa,nguyen2020dataset,nguyen2021dataset,wang2022cafe,mtt}), we propose applying a \textit{deep generative prior} to our distillation process by means of distilling into the latent space of a pre-trained generative model.
We find that such a prior greatly increases the generalization of the synthetic dataset to architectures other than the ``backbone'' model used in the distillation process (the lack of such generalization being one of the largest limitations of previous methods).
Furthermore, our new parametrization facilitates the distillation of even larger-resolution synthetic data devoid of the high-frequency noise induced by previous distillation methods.
Lastly, the deep generative prior acts as a plug-and-play addition to any existing and future methods of dataset distillation.

\looseness=-1
Concretely, we consider a deep generative model $G$ that outputs samples $G(z)$ given latent vector $z$ (\eg a GAN). At distillation time, we parameterize the small synthetic dataset $\mathcal{S}$ as \begin{equation}
    \mathcal{S} \triangleq \{G(z) \colon z \in \mathcal{Z}\},
\end{equation}
where $\mathcal{Z}$ is a set of latent vectors. Since $G$ is fully differentiable, we can optimize $\mathcal{Z}$ \wrt any distillation objective, such as $\mathcal{L}_{\dc{}}$, $\mathcal{L}_{\dm{}}$ or $\mathcal{L}_{\mtt{}}$.
Please see \refalg{main} for a complete write-up of our method.

\begin{table*}[t]
\centering
\resizebox{\textwidth}{!}{%
\begin{tabular}{lccccccccccc}
Distil. Alg.                                              & Distil. Space & ImNet-A            & ImNet-B            & ImNet-C            & ImNet-D            & ImNet-E            & ImNette            & ImWoof    & ImNet-Birds & ImNet-Fruits       & ImNet-Cats         \\\midrule
                                                          & Pixel         & 33.4{\color{gray}$\pm$1.5}          & 34.0{\color{gray}$\pm$3.4}          & 31.4{\color{gray}$\pm$3.4}          & 27.7{\color{gray}$\pm$2.7}          & 24.9{\color{gray}$\pm$1.8}          & 24.1{\color{gray}$\pm$1.8}          & 16.0{\color{gray}$\pm$1.2} & 25.5{\color{gray}$\pm$3.0}   & 18.3{\color{gray}$\pm$2.3}          & 18.7{\color{gray}$\pm$1.5}          \\
\multirow{-2}{*}{\mtt{} \cite{mtt}}                       & \texttt{GLaD} (Ours)    & \textbf{39.9{\color{gray}$\pm$1.2}} & \textbf{39.4{\color{gray}$\pm$1.3}} & \textbf{34.9{\color{gray}$\pm$1.1}}          & \textbf{30.4{\color{gray}$\pm$1.5}}          & \textbf{29.0{\color{gray}$\pm$1.1}} & \textbf{30.4{\color{gray}$\pm$1.5}} & \textbf{17.1{\color{gray}$\pm$1.1}} & \textbf{28.2{\color{gray}$\pm$1.1}}   & \textbf{21.1{\color{gray}$\pm$1.2}}          & \textbf{19.6{\color{gray}$\pm$1.2}}          \\\midrule
\cellcolor[HTML]{FFFFFF}                                  & Pixel         & 38.7{\color{gray}$\pm$4.2}          & 38.7{\color{gray}$\pm$1.0}          & 33.3{\color{gray}$\pm$1.9}          & 26.4{\color{gray}$\pm$1.1}          & 27.4{\color{gray}$\pm$0.9}          & 28.2{\color{gray}$\pm$1.4}          & 17.4{\color{gray}$\pm$1.2} & 28.5{\color{gray}$\pm$1.4}   & 20.4{\color{gray}$\pm$1.5}          & 19.8{\color{gray}$\pm$0.9}          \\
\multirow{-2}{*}{\cellcolor[HTML]{FFFFFF}\dc{} \cite{dc}} & \texttt{GLaD} (Ours)    & \textbf{41.8{\color{gray}$\pm$1.7}}          & \textbf{42.1{\color{gray}$\pm$1.2}} & \textbf{35.8{\color{gray}$\pm$1.4}}          & \textbf{28.0{\color{gray}$\pm$0.8}} & \textbf{29.3{\color{gray}$\pm$1.3}}          & \textbf{31.0{\color{gray}$\pm$1.6}}          & \textbf{17.8{\color{gray}$\pm$1.1}} & \textbf{29.1{\color{gray}$\pm$1.0}}   & \textbf{22.3{\color{gray}$\pm$1.6}} & \textbf{21.2{\color{gray}$\pm$1.4}} \\\midrule
\cellcolor[HTML]{FFFFFF}                                  & Pixel         & 27.2{\color{gray}$\pm$1.2}          & 24.4{\color{gray}$\pm$1.1}          & 23.0{\color{gray}$\pm$1.4}          & 18.4{\color{gray}$\pm$0.7}          & 17.7{\color{gray}$\pm$0.9}          & 20.6{\color{gray}$\pm$0.7}          & 14.5{\color{gray}$\pm$0.9} & 17.8{\color{gray}$\pm$0.8}   & 14.5{\color{gray}$\pm$1.1}          & 14.0{\color{gray}$\pm$1.1}          \\
\multirow{-2}{*}{\cellcolor[HTML]{FFFFFF}\dm{} \cite{dm}} & \texttt{GLaD} (Ours)    & \textbf{31.6{\color{gray}$\pm$1.4}} & \textbf{31.3{\color{gray}$\pm$3.9}} & \textbf{26.9{\color{gray}$\pm$1.2}} & \textbf{21.5{\color{gray}$\pm$1.0}} & \textbf{20.4{\color{gray}$\pm$0.8}} & \textbf{21.9{\color{gray}$\pm$1.1} }         & \textbf{15.2{\color{gray}$\pm$0.9}} & \textbf{18.2{\color{gray}$\pm$1.0}}   & \textbf{20.4{\color{gray}$\pm$1.6}} & \textbf{16.1{\color{gray}$\pm$0.7}}
\end{tabular}
}
\caption{
\textbf{ImageNet (128$\times$128) Performance on Unseen Architectures}. These results come from training AlexNet, VGG11, ResNet18, and a Vision Transformer on our synthetic datasets (that were distilled using a ConvNet) and averaging their performances on the real validation sets. Applying the deep generative prior by distilling into F-space rather than pixel space significantly improves the cross-architecture generalization of all methods across all sampled datasets.
}
\label{tab:template}
\end{table*}

\begin{table}[]
\centering
\resizebox{\linewidth}{!}{%

\begin{tabular}{lcccccc}
Distil. Alg.                                    & Distil. Space  & ImNet-A                    & ImNet-B                    & ImNet-C                    & ImNet-D                    & ImNet-E                    \\ \midrule
                                                & Pixel          & 52.3{\color{gray}$\pm$0.7} & 45.1{\color{gray}$\pm$8.3} & 40.1{\color{gray}$\pm$7.6} & 36.1{\color{gray}$\pm$0.4} & 38.1{\color{gray}$\pm$0.4} \\ 
\multirow{-2}{*}{\cellcolor[HTML]{FFFFFF}\dc{}} & \glad{} (Ours) & \textbf{53.1{\color{gray}$\pm$1.4}} & \textbf{50.1{\color{gray}$\pm$0.6}} & \textbf{48.9{\color{gray}$\pm$1.1}} & \textbf{38.9{\color{gray}$\pm$1.0}} & \textbf{38.4{\color{gray}$\pm$0.7}} \\ \midrule
\cellcolor[HTML]{FFFFFF}                        & Pixel          & 52.6{\color{gray}$\pm$0.4} & 50.6{\color{gray}$\pm$0.5} & 47.5{\color{gray}$\pm$0.7} & 35.4{\color{gray}$\pm$0.4} & 36.0{\color{gray}$\pm$0.5} \\
\multirow{-2}{*}{\cellcolor[HTML]{FFFFFF}\dm{}} & \glad{} (Ours) & \textbf{52.8{\color{gray}$\pm$1.0}} & \textbf{51.3{\color{gray}$\pm$0.6}} & \textbf{49.7{\color{gray}$\pm$0.4}} & \textbf{36.4{\color{gray}$\pm$0.4}} & \textbf{38.6{\color{gray}$\pm$0.7}}
\end{tabular}
}

\caption{Performance with 10 images/class.
}
\label{tab:10ipc}
\end{table}
\begin{table}
\centering
\resizebox{\columnwidth}{!}{%

\setlength{\tabcolsep}{3pt}
\begin{tabular}{llccccc}
\multirow{2}{*}[-0.2em]{\shortstack{Distillation\\[0.1em]Method}}                     & \hspace{1.5em}\multirow{2}{*}[-0.2em]{\shortstack{Distillation\\[0.1em]Space}} &    \multicolumn{5}{c}{\textbf{Unseen} Evaluation Architecture}   \\ \cmidrule{3-7}
&&AlexNet       & ResNet18      & VGG11         & ViT           & Average  \\
\midrule
\multirow{3}{*}{\mtt{} \cite{mtt}} & Pixel            & 26.8{\color{gray}$\pm$0.6}         & 23.4{\color{gray}$\pm$1.3}          & 24.9{\color{gray}$\pm$0.8}          & 21.2{\color{gray}$\pm$0.4}          & 24.1{\color{gray}$\pm$0.8}          \\
                                   & \texttt{GLaD} (Rand G)    & \cellcolor{blue!10}27.4{\color{gray}$\pm$0.3}          & \cellcolor{blue!10}30.1{\color{gray}$\pm$1.2}          & \cellcolor{blue!10}29.0{\color{gray}$\pm$0.8}         & \cellcolor{blue!10}21.9{\color{gray}$\pm$0.3}          & \cellcolor{blue!10}27.1{\color{gray}$\pm$0.7}          \\
                                   & \texttt{GLaD} (Trained G) & \cellcolor{blue!30}\textbf{27.9{\color{gray}$\pm$0.6}} & \cellcolor{blue!30}\textbf{30.2{\color{gray}$\pm$0.6}} & \cellcolor{blue!30}\textbf{31.3{\color{gray}$\pm$0.7}} & \cellcolor{blue!30}\textbf{22.7{\color{gray}$\pm$0.4}} & \cellcolor{blue!30}\textbf{28.0{\color{gray}$\pm$0.6}} \\ \midrule
\multirow{3}{*}{\dc{} \cite{dc}}   & Pixel            & 25.9{\color{gray}$\pm$0.2}          & 27.3{\color{gray}$\pm$0.5}          & 28.0{\color{gray}$\pm$0.5}          & \cellcolor{blue!10}22.9{\color{gray}$\pm$0.3}          & 26.0{\color{gray}$\pm$0.4}          \\
                                   & \texttt{GLaD} (Rand G)    & \cellcolor{blue!30}\textbf{30.1{\color{gray}$\pm$0.5}} & \cellcolor{blue!10}27.3{\color{gray}$\pm$0.2}         & \cellcolor{blue!10}28.0{\color{gray}$\pm$0.9}         & 21.2{\color{gray}$\pm$0.6}          & \cellcolor{blue!30}\textbf{26.6{\color{gray}$\pm$0.5}} \\
                                   & \texttt{GLaD} (Trained G) & \cellcolor{blue!10}26.0{\color{gray}$\pm$0.7}          & \cellcolor{blue!30}\textbf{27.6{\color{gray}$\pm$0.6}} & \cellcolor{blue!30}\textbf{28.2{\color{gray}$\pm$0.6}} & \cellcolor{blue!30}\textbf{23.4{\color{gray}$\pm$0.2}} & \cellcolor{blue!10}26.3{\color{gray}$\pm$0.5}          \\ \midrule
\multirow{3}{*}{\dm{} \cite{dm}}   & Pixel            & 22.9{\color{gray}$\pm$0.2}          & \cellcolor{blue!10}22.2{\color{gray}$\pm$0.7}          & 23.8{\color{gray}$\pm$0.5}          & 21.3{\color{gray}$\pm$0.5}          & 22.6{\color{gray}$\pm$0.5}         \\
                                   & \texttt{GLaD} (Rand G)    & \cellcolor{blue!10}23.7{\color{gray}$\pm$0.3}         & 21.7{\color{gray}$\pm$1.0}         & \cellcolor{blue!10}24.3{\color{gray}$\pm$0.8}         & \cellcolor{blue!10}21.4{\color{gray}$\pm$0.2}         & \cellcolor{blue!10}22.8{\color{gray}$\pm$0.6}         \\
                                   & \texttt{GLaD} (Trained G) & \cellcolor{blue!30}\textbf{25.1{\color{gray}$\pm$0.5}} & \cellcolor{blue!30}\textbf{22.5{\color{gray}$\pm$0.7}} & \cellcolor{blue!30}\textbf{24.8{\color{gray}$\pm$0.8}}& \cellcolor{blue!30}\textbf{23.0{\color{gray}$\pm$0.1}} & \cellcolor{blue!30}\textbf{23.8{\color{gray}$\pm$0.5}}
\end{tabular}
}
\caption{
\cellcolor{blue!30}\textbf{CIFAR-10 Performance on Unseen Architectures}. Unlike the high-resolution data, we only see a large improvement to \mtt{} and a moderate improvement to \dm{}. Interestingly, we also see that distilling into the latent space of an \textit{un-trained} generator still yields results on-par or better than pixel-space distillation.
}

\label{tab:cifar}
\end{table}

\subsection{Choosing a Generative Model and Latent Space}

For a flexible and effective parameterization, we fill the role of our deep generative prior with the recently proposed StyleGAN-XL \cite{sgxl}, a modified version of StyleGAN3 \cite{sg3}. StyleGAN generators can not only output high-fidelity images \cite{stylegan2ada}, but also (1) provide multiple flexible latent spaces for parameterizing images \cite{abdal2019image2stylegan,parmar2022sam} and (2) inherently impose diverse and interesting priors via architecture (even at random initialization) \cite{baradad2021learning}. In our experiments, this allows us to  probe the effects of choosing latent spaces from different layers and using generators trained on out-of-distribution datasets or even at random initialization.

\looseness=-1
Distilling into the latent space of StyleGAN-XL can be thought of as a pseudo-inversion task. 
However, we found that distilling into even extended W\textsuperscript{+} latent space\footnote{Here W space refers to the output space of StyleGAN-XL's MLP ``mapping'' network, and W\textsuperscript{+} allows for different W space vectors at different layers.}\cite{abdal2019image2stylegan}, the most flexible of the traditional StyleGAN inversion spaces, was too restrictive for our objective (see \reffig{fig:layers}). 
It limits the synthetic images to be realistic, but, unlike real image inversion, these images are optimized for the distillation task and do not require realism.  %
Prior inversion works propose the ``F$n$'' spaces in StyleGAN as an alternative that allows images to be more diverse and flexible  \cite{zhu2021barbershop,parmar2022sam}.
As such, we choose to distill into these ``F$n$'' spaces, which means optimizing the $n$-th hidden layer of StyleGAN-XL's ``synthesis'' network's latent representation along with all subsequent W\textsuperscript{+} modulation codes.
Note that in this work, $z$ and $\mathcal{Z}$ refer to the concatenation of the F$n$ feature map with the W\textsuperscript{+} codes, \textbf{not} the traditional StyleGAN $z$-space. 
Please see the \supp for details of the  StyleGAN-XL architecture and the ``F$n$'' spaces.

\looseness=-1
Since these ``F$n$'' spaces are from intermediate layers and do not have associated prior distributions, we initialize our latent $z$ vectors using the empirical distribution of latent vectors of the corresponding classes. With access to the earlier layers of $G$, this can be easily computed. Please see the \supp for details of our initialization scheme.

\begin{table}[t]
\centering
\resizebox{\columnwidth}{!}{%

\begin{tabular}{lccccc}
Distillation Space                        & ImNet-A   & ImNet-B   & ImNet-C   & ImNet-D   & ImNet-E   \\ \midrule
Pixel                                & 38.3{\color{gray}$\pm$4.7} & 32.8{\color{gray}$\pm$4.1} & 27.6{\color{gray}$\pm$3.3} & 25.5{\color{gray}$\pm$1.2} & 23.5{\color{gray}$\pm$2.4} \\
\texttt{GLaD} (ImageNet G)                            & 37.4{\color{gray}$\pm$5.5} & \cellcolor{blue!30}\textbf{41.5{\color{gray}$\pm$1.2}} & \cellcolor{blue!30}\textbf{35.7{\color{gray}$\pm$4.0}} & \cellcolor{blue!20}27.9{\color{gray}$\pm$1.0} & 29.3{\color{gray}$\pm$1.2}\\
\texttt{GLaD} (Random G)                        & \cellcolor{blue!30}\textbf{39.3{\color{gray}$\pm$2.0}} & \cellcolor{blue!20}40.3{\color{gray}$\pm$1.7} & 35.0{\color{gray}$\pm$1.7} & \cellcolor{blue!20}27.9{\color{gray}$\pm$1.4} & \cellcolor{blue!30}\textbf{29.8{\color{gray}$\pm$1.4}} \\
\cellcolor[HTML]{FFFFFF}\texttt{GLaD} (Pokémon G) & \cellcolor{blue!20}39.1{\color{gray}$\pm$2.0} & 39.4{\color{gray}$\pm$1.5} & \cellcolor{blue!20}35.3{\color{gray}$\pm$1.3} & \cellcolor{blue!30}\textbf{28.0{\color{gray}$\pm$1.2}} & \cellcolor{blue!20}29.5{\color{gray}$\pm$1.3} \\
\texttt{GLaD} (FFHQ G)                          & 38.3{\color{gray}$\pm$5.2} & 40.2{\color{gray}$\pm$1.1} & 34.9{\color{gray}$\pm$1.1} & 27.2{\color{gray}$\pm$0.9} & 29.4{\color{gray}$\pm$2.1 }

\end{tabular}
}
\caption{
Higher-resolution (256$\times$256) datasets distilled (using \dc{}) into the latent space of a randomly-initialized generator as well as generators on Pokémon or FFQH still realize significant improvements in cross-architecture generalization over their pixel-space counterparts.
}

\label{tab:other}
\end{table}

\subsection{Memory Saving via Checkpointing}
\lblsec{MC}
As the forward pass through modern deep generative models usually requires copious amounts of GPU VRAM, our method (if implemented naively) becomes difficult to run on a limited number of GPUs.
To circumvent this issue, we employ a technique inspired by gradient checkpointing \citep{chen2016training}.
At each distillation iteration, we first obtain our synthetic images $\mathcal{S} = G(\mathcal{Z})$ \textit{without} tracking any gradients.
We then calculate our distillation loss $\mathcal{L}$, compute the gradient of this loss with respect to our synthetic images ($\partial\mathcal{L}/\partial\mathcal{S}$), and delete the computation graph used to compute $\mathcal{L}$ and its gradient.
To compute $\partial \mathcal{L} / \partial{\mathcal{Z}}$, we \textit{re}-compute the forward pass through $G$, $\mathcal{S} = G(\mathcal{Z})$, this time tracking gradients such that we know $\partial \mathcal{S} / \partial \mathcal{Z}$. 
From here, application of the chain rule gives us $\partial \mathcal{L}/ \partial \mathcal{Z} = (\partial\mathcal{L}/\partial\mathcal{S})(\partial\mathcal{S}/\partial \mathcal{Z})$ which we use to update the latent codes for our synthetic data.
For example, with $128\times128$-resolution StyleGAN-XL, this memory-saving trick allows us to save nearly $2$GB memory per synthetic image with F$0$ space.

\begin{figure*}[tp]
    \centering
    \setlength{\tabcolsep}{1pt}
    \begin{tabular}{ccc}
        ImageNet-A with \mtt{} \cite{mtt} & ImageNet-B with \dc{} \cite{dc} & ImageNet-C with \dm{} \cite{dm}\\\vspace{-0.1cm}
        
        \includegraphics[width=0.32\linewidth]{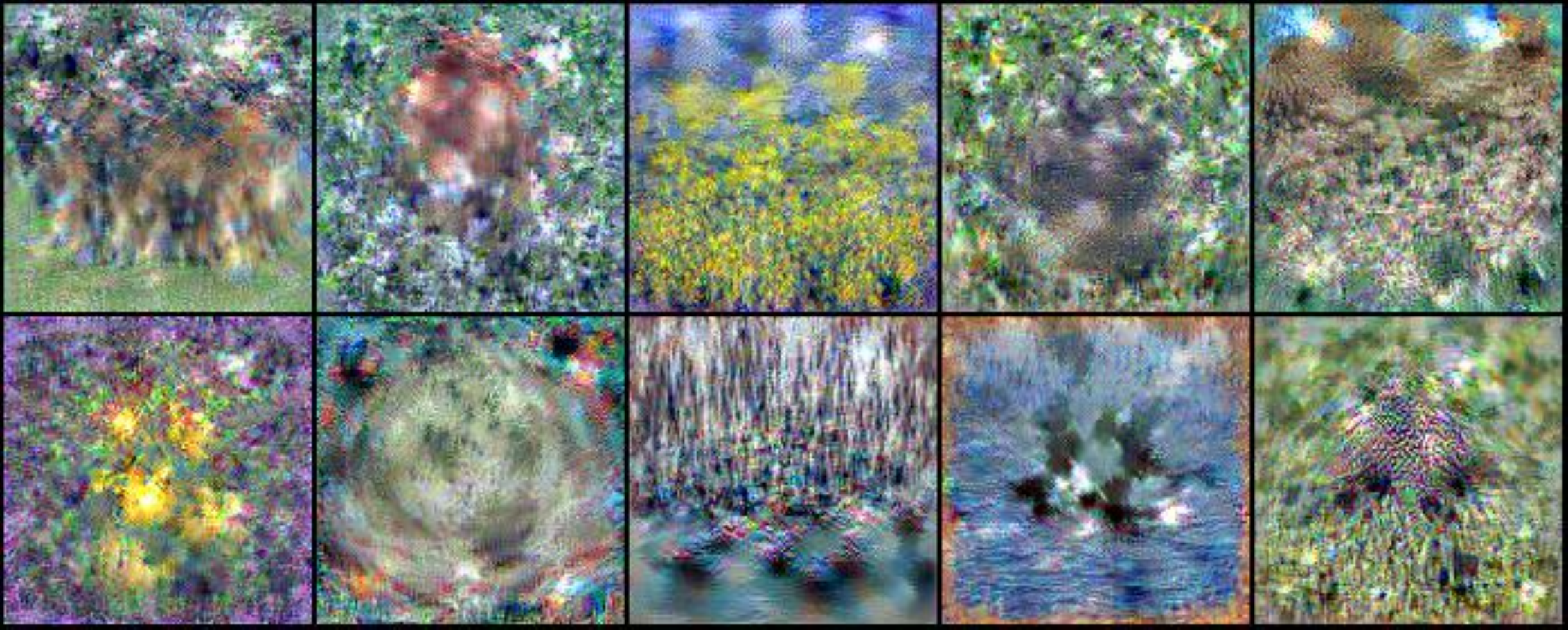} &  
        \includegraphics[width=0.32\linewidth]{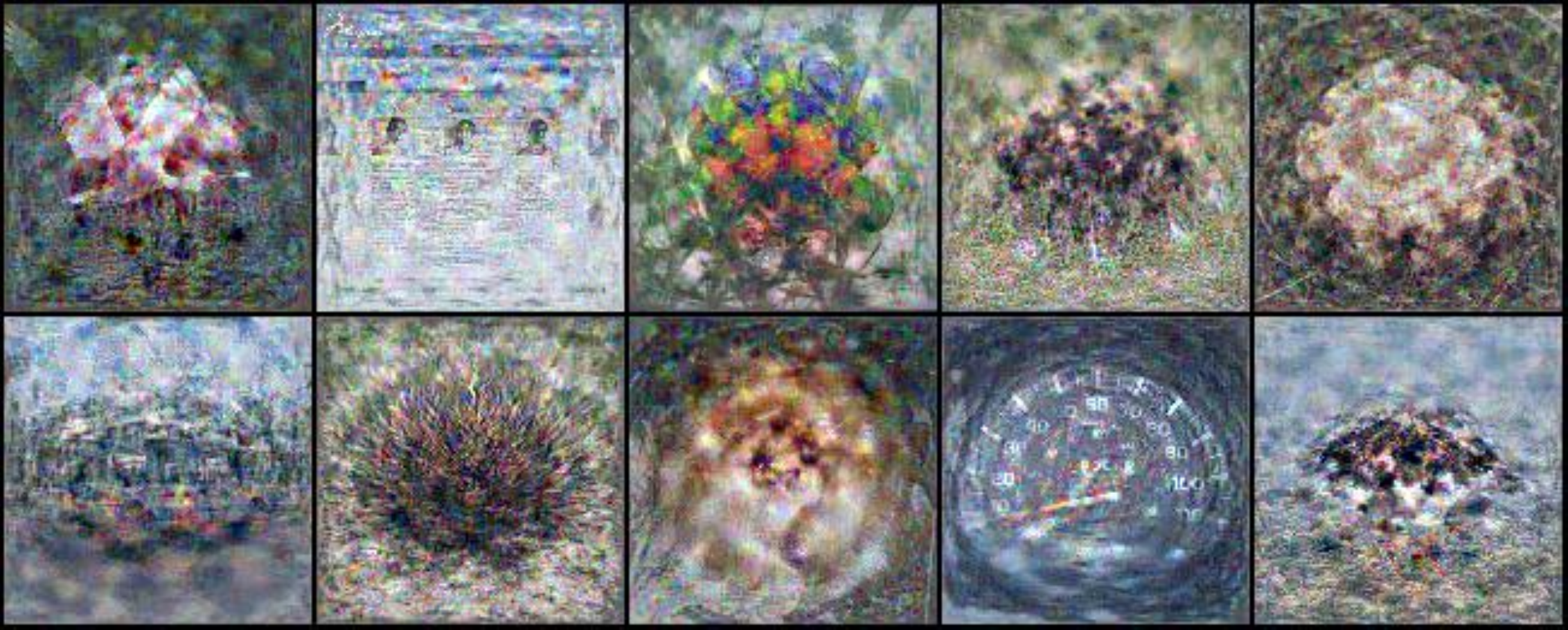} &
        \includegraphics[width=0.32\linewidth]{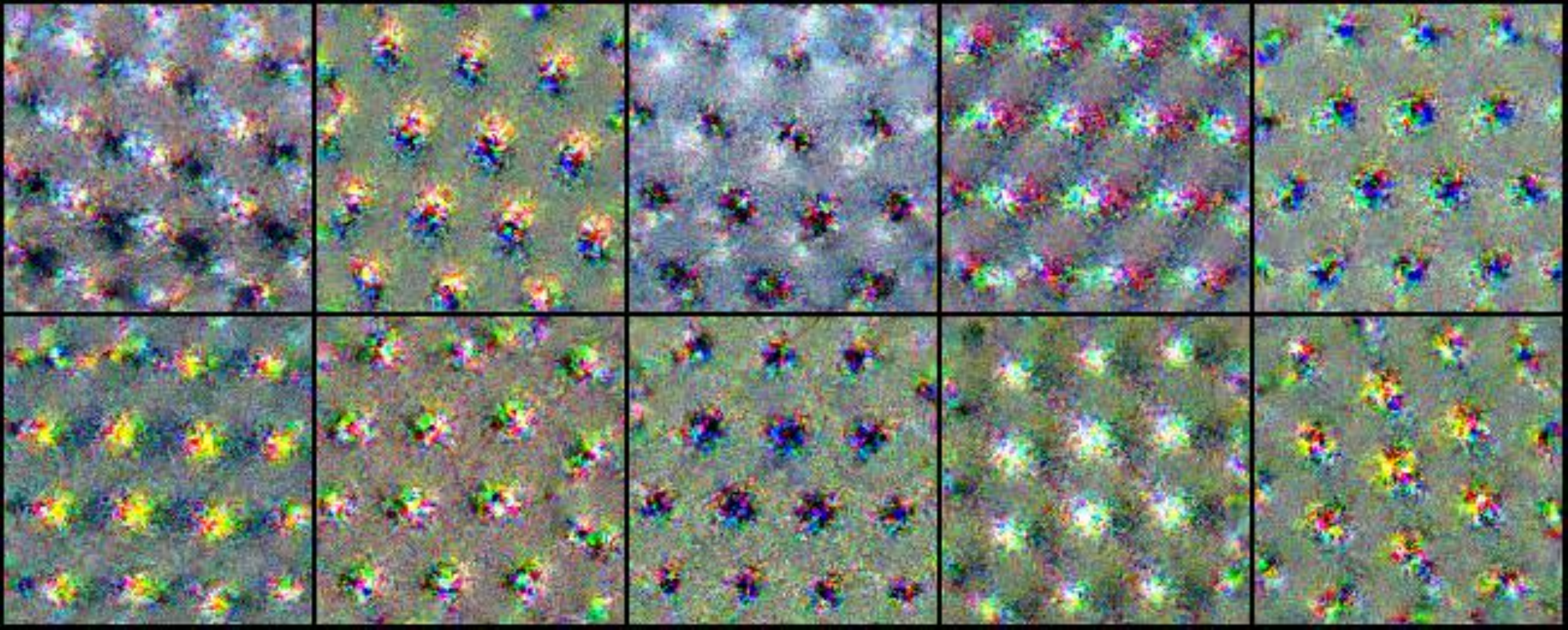}\\
        \smaller{Pixel-Space: Cross-Arch 33.4}\% & \smaller{Pixel-Space: Cross-Arch: 38.7\%} & \smaller{Pixel-Space: Cross-Arch 23.0\%}\\\vspace{-0.1cm}
        \includegraphics[width=0.32\linewidth]{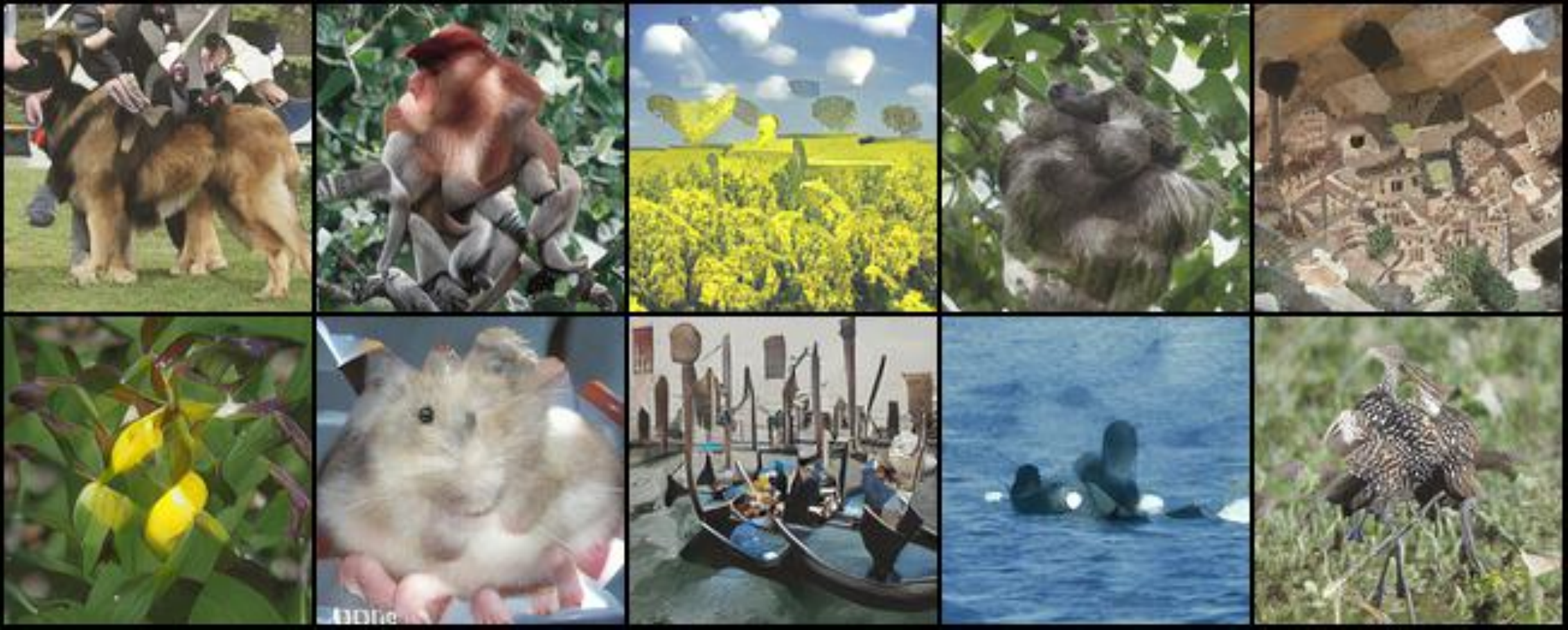} &  
        \includegraphics[width=0.32\linewidth]{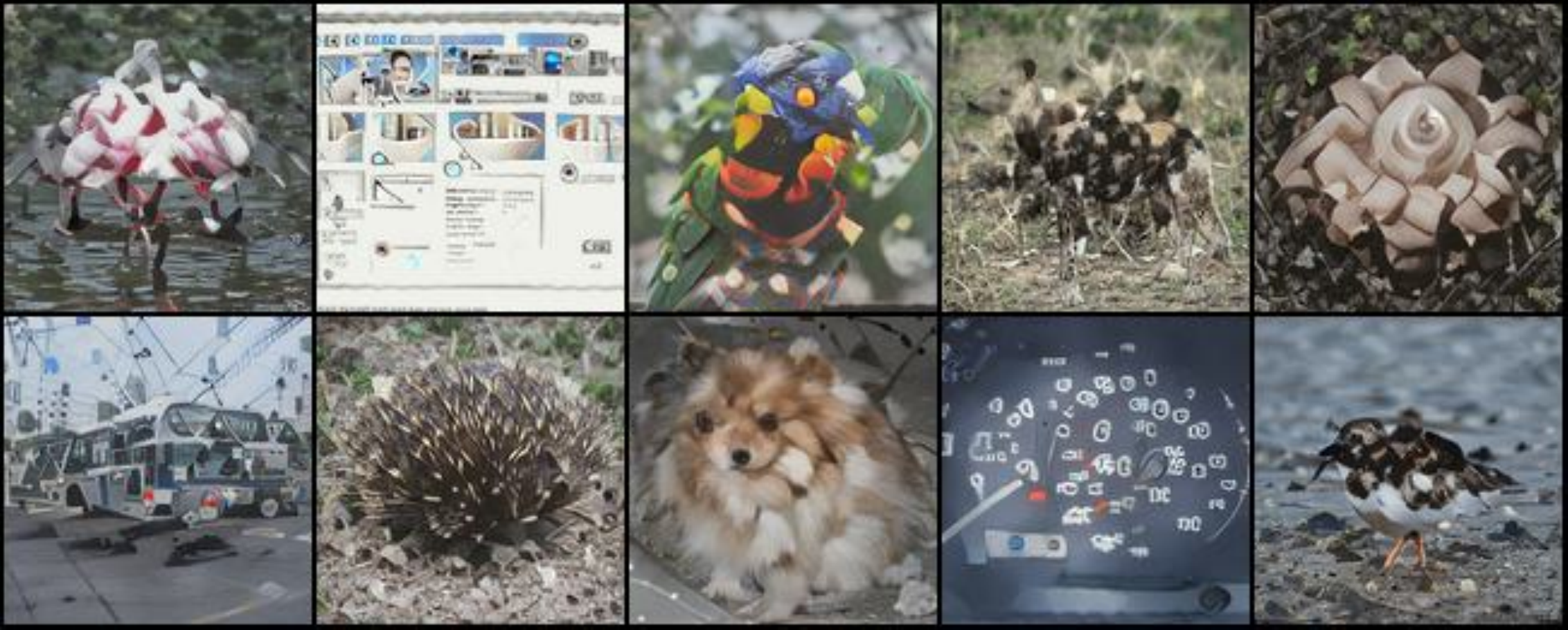} &
        \includegraphics[width=0.32\linewidth]{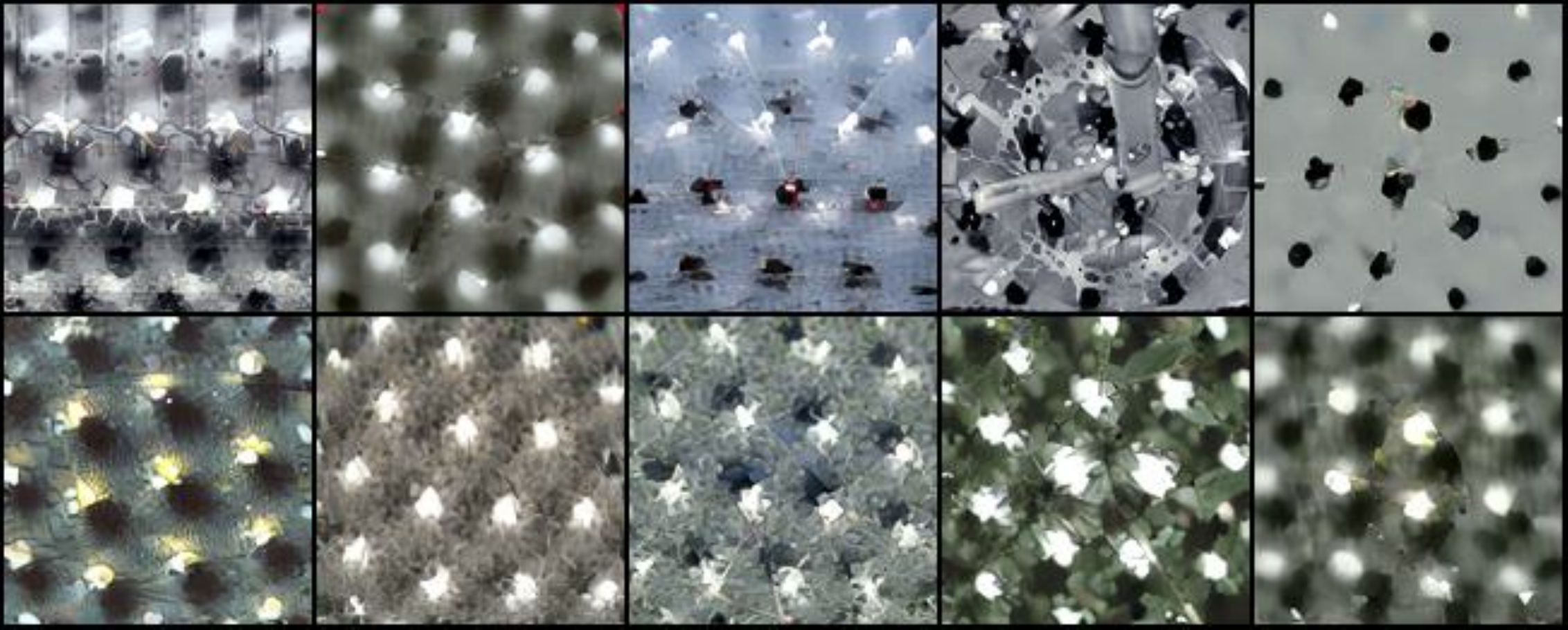}\\
        \smaller{\glad{}: Cross-Arch \textbf{39.9\%}} & \smaller{\glad{}: Cross-Arch \textbf{42.1\%}} & \smaller{\glad{}: Cross-Arch \textbf{26.9\%}}
    \end{tabular}
    \vspace{-10pt}
    \caption{Across all sampled ImageNet subsets and all distillation methods, the addition of the generative prior (\textbf{bottom}) offers significantly better cross-architecture generalization than pixel-space distillation (\textbf{top}).}
    \label{fig:template}
\end{figure*}

\begin{figure}
     \centering
     \begin{subfigure}[b]{\linewidth}
         \centering
         \includegraphics[width=\linewidth]{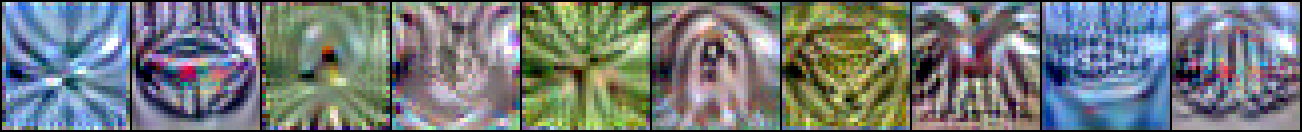}
         \caption{Pixel-Space Distillation; 24.1\% Cross Arch. Acc.}
         \label{fig:y equals x}
     \end{subfigure}
     \hfill
     \begin{subfigure}[b]{\linewidth}
         \centering
         \includegraphics[width=\linewidth]{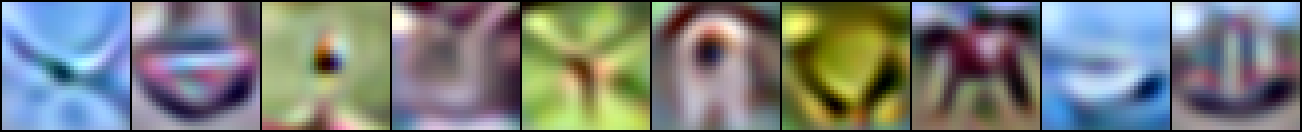}
         \caption{F2-Space Distillation, Random Generator; 27.1\% Cross Arch. Acc.}
         \label{fig:three sin x}
     \end{subfigure}
     \hfill
     \begin{subfigure}[b]{\linewidth}
         \centering
         \includegraphics[width=\linewidth]{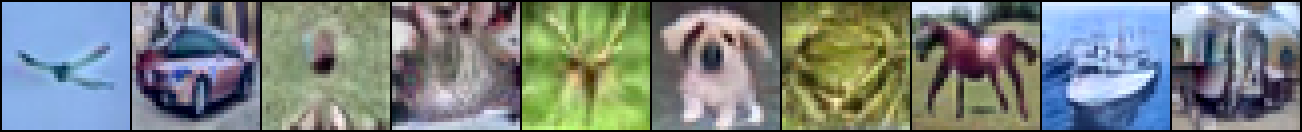}
         \caption{F2-Space Distillation, Pre-Trained Generator; 28.0\% Cross Arch. Acc.}
         \label{fig:five over x}
     \end{subfigure}
    \vspace{-15pt}
    \caption{CIFAR-10 distillations using \mtt{}. The images distilled with both the random and pre-trained generators lead to better cross-architecture generalization than those distilled into pixel space.}
        \label{fig:three graphs}
\end{figure}

\section{Experiments}\lblsec{expr}
\looseness=-1
We evaluate our method (\texttt{GLaD}) for distilling CIFAR-10 \cite{cifar10} and 10-class subsets of ImageNet 1k \cite{deng2009imagenet}, utilizing StyleGAN-XL \cite{sgxl} generators trained on these datasets (obtained from the official released model checkpoints). 

The code for our experiments is based on the open-source repositories for \dc{}, \dm{}, and \mtt{} and will be released upon publication. 
For each method, we integrate the deep generative prior directly into the existing code base.
For results with and without the generative prior, we use the same set of hyper-parameters ($N$, $M$, $T^+$, \#iterations, etc.) to ensure a fair comparison. For each method, we choose an F$n$ space and use it for all datasets. %
Please see the \supp for experiment details.
\begin{figure}[tp]
    \setlength\tabcolsep{0.5pt}
    \centering
    \begin{tabular}{rccccccc}
        \rotatebox[origin=c]{90}{\smaller{Init.}} & \hspace{1.5pt}
        \includegraphics[align=c,width=.13\linewidth]{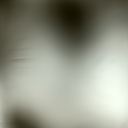} &
        \includegraphics[align=c,width=.13\linewidth]{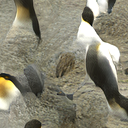} &
        \includegraphics[align=c,width=.13\linewidth]{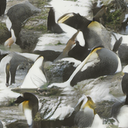} &
        \includegraphics[align=c,width=.13\linewidth]{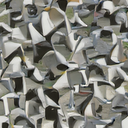} &
        \includegraphics[align=c,width=.13\linewidth]{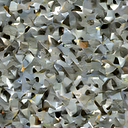} &
        \includegraphics[align=c,width=.13\linewidth]{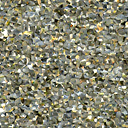} &
        \includegraphics[align=c,width=.13\linewidth]{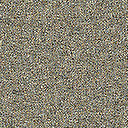}\\[1.05em]
        \rotatebox[origin=c]{90}{\smaller{Final}} & \hspace{1.5pt}
        \includegraphics[align=c,width=.13\linewidth]{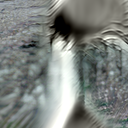} &
        \includegraphics[align=c,width=.13\linewidth]{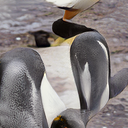} &
        \includegraphics[align=c,width=.13\linewidth]{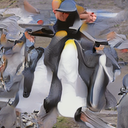} &
        \includegraphics[align=c,width=.13\linewidth]{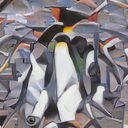} &
        \includegraphics[align=c,width=.13\linewidth]{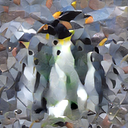} &
        \includegraphics[align=c,width=.13\linewidth]{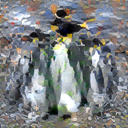} &
        \includegraphics[align=c,width=.13\linewidth]{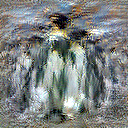}\\[1.2em]
        &F0 & F4 & F8 & F12 & F16 & F20 & F24
    \end{tabular}
    \caption{Replacing the feed-forward initialization of the F latents with random noise has drastically different effects in the initial image (\textbf{top}) depending on which layer is used. In the optimized images (\textbf{bottom}), we see artifacts with varying levels of granularity depending on the layer.}
    \label{fig:randf}
\end{figure}

\myparagraph{Datasets.}
For low-resolution data, we apply our method to CIFAR-10 \cite{cifar10}. 
For higher-resolution data, we use \textit{subsets} of ImageNet-1k \cite{deng2009imagenet}. Previous dataset distillation work \cite{mtt} introduced several subsets based largely on categories and aesthetics, including birds, fruits, and cats. 
Two other conventional subsets are ImageNette and ImageWoof \cite{imagenette}. We also introduce several new 10-class subsets based on the evaluation performance of a ResNet-50 model that has been pre-trained on ImageNet. 
In this work, ``ImageNet-A'' consists of the top-10 classes with  ``ImageNet-B'' consisting of the next 10 and so on for ``ImageNet-C,'' ``ImageNet-D,'' and ``ImageNet-E.'' The classes composing all the subsets can be found in the \supp.
\begin{figure}[t]
    \centering
    \small
    \setlength\tabcolsep{1.5pt}
    \begin{tabular}{cc}
    \includegraphics[width=0.48\linewidth]{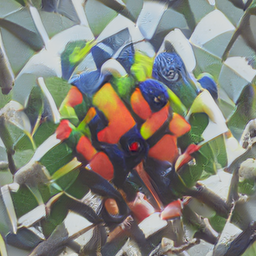} & 
    \includegraphics[width=0.48\linewidth]{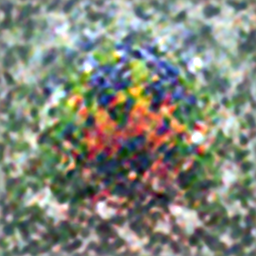} \\[-0.3ex]
    ImageNet GAN & Random GAN \\[0.6ex]
    \includegraphics[width=0.48\linewidth]{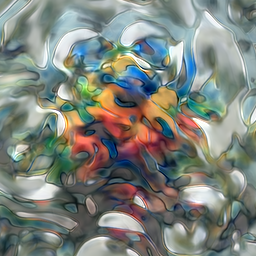}& \includegraphics[width=0.48\linewidth]{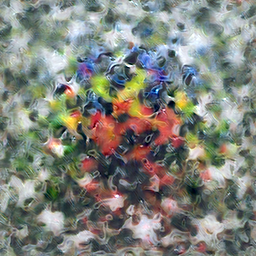}\\[-0.3ex]
    Pokémon GAN & FFHQ GAN
    \end{tabular}
    \caption{\looseness=-1
    Distilling into the latent space of generators that have \textbf{not} been trained on the relevant data still produces good images. Here we show a ``lorikeet'' distilled using a randomly initialized generator as well as generators trained on ImageNet, Pokémon, and FFHQ.}
    \label{fig:other_gans}
\end{figure}

\myparagraph{Evaluation Protocol.}
After distilling our synthetic datasets with their respective algorithm, we then evaluate them on a set of unseen architectures.
To evaluate a synthetic dataset on a given architecture, we train a network from scratch on the distilled dataset and then evaluate it on the validation set.

The training regiment is the same for all networks and datasets: SGD with momentum, $\ell_2$ weight decay, and 500 epochs of linear warm-up followed by another 500 of cosine decay. 
An appropriate (fixed) starting learning rate is used for each architecture, and the final validation set evaluation is done using the exponential moving average of the model's weights. 
This process is repeated 5 times, and the mean validation accuracy $\pm$ 1 standard deviation is reported. Further details can be found in the \supp.

\myparagraph{Network Architectures.}
As with prior dataset distillation works \cite{dd,dc,dsa,dm, nguyen2020dataset,nguyen2021dataset,wang2022cafe,tesla,frepo}, we use the ConvNet \cite{gidaris2018dynamic} architecture as our backbone network. 
A Depth-$n$ ConvNet consists of $n$ blocks followed by a fully-connected layer where each block consists of a $3\times 3$ convolutional layer with 128 filters, instance normalization \cite{ulyanov2016instance}, ReLU non-linearity, and $2\times 2$ average pooling with stride 2.

\looseness=-1
We used two sets of models for our cross-architecture generalization experiments.
For our CIFAR experiments, we use the AlexNet \cite{alexnet}, VGG-11 \cite{vgg}, and ResNet-18 \cite{resnet} included with the codebase of \dc{}/\dsa{}/\dm{}/\mtt{} along with a Vision Transformer \cite{vit} from the DC-BENCH \cite{cui2022dc} repository.
For our experiments on higher-resolution data, we use slightly modified versions of the networks to accommodate the larger images.

Our experiments are enumerated below to highlight the contributions of our proposed method.

\subsection{Finding a Suitable Latent Space}
\looseness=-1
Given the depth of StyleGAN-XL \cite{sgxl}, there are \textit{many} possible latent spaces that \texttt{GLaD} can use to parameterize our synthetic dataset. 
To find the best latent space for cross-architecture generalization, we experiment with several ImageNet subsets. 
In \reffig{fig:layers}, we see the final distilled images for the ``flamingo'' class. 
Earlier latent spaces enforce a stronger prior on the distilled image, while later latent spaces offer more flexibility for the optimization to fit to the distillation objective (\mtt{}, \dc{}, or \dm{}). 
Examining \reffig{fig:layers}, we see that the images distilled into F12 space result in the best cross-architecture generalization for \mtt{}. Through analogous experiments, we found F16 to be optimal for \dc{} and F20 for \dm{}.

\begin{figure}[t]
    \centering
    \setlength\tabcolsep{1.5pt}
    \small
    \begin{tabular}{ccc}
    \includegraphics[width=0.31\linewidth]{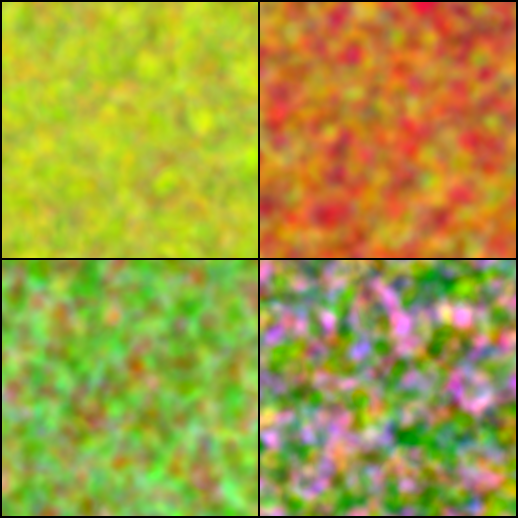} & 
    \includegraphics[width=0.31\linewidth]{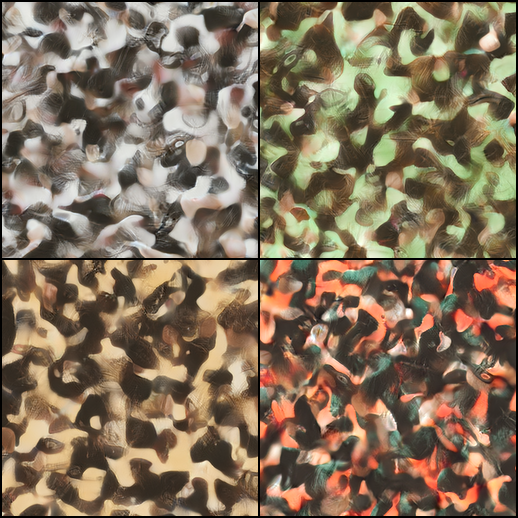} &
    \includegraphics[width=0.31\linewidth]{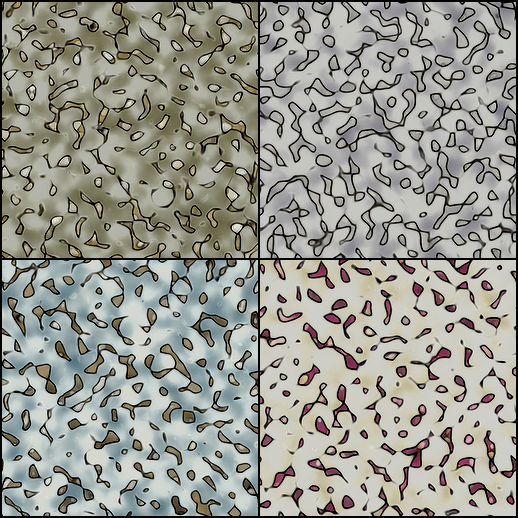}\\[-0.3ex]
    Random GAN & FFHQ GAN & \pkmn~GAN
    \end{tabular}
    \caption{When initializing the latents with random noise, each generator has a unique type of structure present in the output. The final optimized images (Figure \ref{fig:other_gans}) contain artifacts of this structure.}
    \label{fig:other_inits}
\end{figure}

\subsection{Improving Cross-Architecture Generalization}\lblsec{CrossArch}
Arguably the most lacking point of all previous dataset distillation methods, cross-architecture generalization gives a good understanding of how well the distillation method ``understands'' the classification task rather than simply overfitting to a given architecture.
In Table \ref{tab:template}, we show cross-architecture results for \mtt{}, \dc{}, and \dm{} with and without the deep generative prior. For each method and dataset, a 1 image-per-class synthetic set is distilled using a Depth-5 ConvNet as the ``backbone'' architecture. 
To evaluate cross-architecture generalization, we use the distilled set to train AlexNet \cite{alexnet}, ResNet-18 \cite{resnet}, VGG-11 \cite{vgg}, and ViT-b/16 \cite{vit} from scratch and record the validation accuracy. We record the average validation accuracy across these 4 architectures in Table \ref{tab:template}.

For every tested dataset, \texttt{GLaD}'s addition of the generative prior slightly or significantly improved the cross-architecture generalization of all 3 methods.

\looseness=-1 
In Table \ref{tab:cifar}, we also include cross-architecture results for CIFAR-10. Even on this lower-resolution data, \texttt{GLaD} significantly improves the performance of both original \mtt{} and original \dm{} while only showing marginal gains on \dc{}.

\begin{figure}[tp]
    \centering
    \small
    \includegraphics[width=1.00\linewidth]{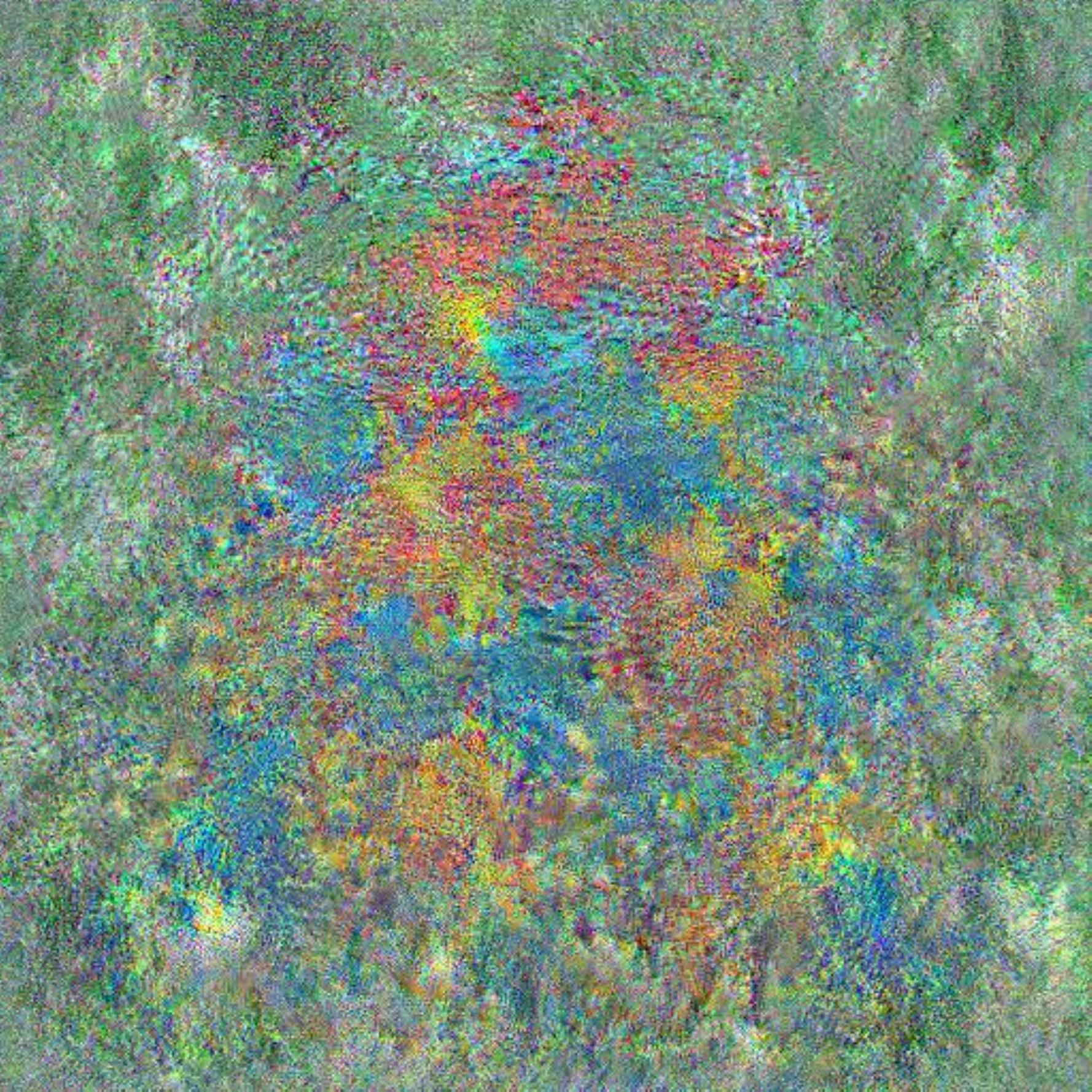}\\[-0.3ex]
    Pixel-Space\vspace{0.2cm}
    \includegraphics[width=1.00\linewidth]{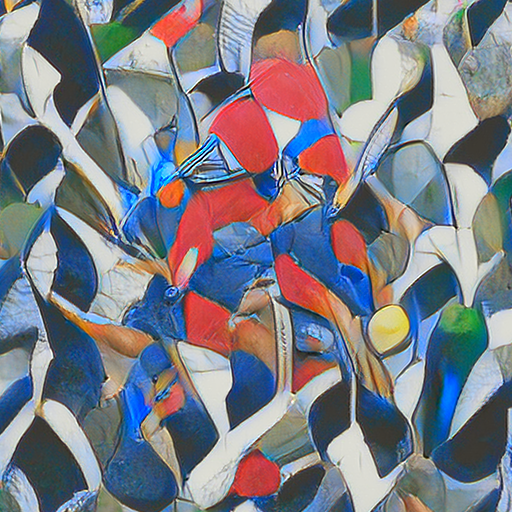}\\[-0.3ex]
    \glad{}
    \caption{Generative Latent Distillation (\glad{}) allows us to distill datasets into high-resolution artistic images (\textbf{bottom}), while high-resolution images distilled into pixel space degenerate into high-frequency patterns (\textbf{top}). This example is a 512$\times$512 ``macaw'' distilled using \mtt{} in pixel space versus \glad{}.}
    \label{fig:512}
\end{figure}

\subsection{Latent Initialization and Generator Choices}
\looseness=-1
In previous experiments, we initialized latents~$z$ (which contain outputs of intermediate layers) via a partial feed-forward pass through the generator.
Here, we experiment with an alternative initialization wherein the feed-forward initialization is replaced with Gaussian noise (with matching mean and variance).
As the StyleGAN generator expects a somewhat coherent representation in the latent space, the images at initialization show interesting artifacts that vary in granularity depending on the latent space used (Figure \ref{fig:randf}).

Using such Gaussian initialization, we found that \texttt{GLaD} can successfully use generators that were trained on \textit{completely} different datasets (such as FFQH \cite{karras2019style} and \pkmn{} \cite{pokemon}) or even generators that have not been trained at all (a-la Deep Image Prior \cite{ulyanov2018deep}). In Table \ref{tab:other}, we compare the pixel-space and standard GAN results to generative latent distillations using generators trained on the FFHQ and Pokémon along with generators that have not been trained at all. These results use \dc{}, and results using \mtt{} and \dm{} can be found in the \supp.

The images distilled using the FFHQ, Pokémon, and random generators still offer cross-architectural generalization improvements over those distilled directly into pixel space, ofter matching or surpassing the results using the standard ImageNet generator.

We also notice that initializing the latents with random Gaussian noise results in aesthetically pleasing images with different ``artistic'' properties based on the generator used (Figure \ref{fig:other_gans}). This trend also extends to even larger images, allowing our method to create class-based digital arts in different styles, such as the ``mosaic'' in Figure \ref{fig:512}.

\section{Discussion and Limitations}
\label{sec:conclusion}
\looseness=-1
In this work, we have proposed leveraging a generative prior to dataset distillation (\glad{}). 
By applying our deep generative prior, we introduce a new method that significantly improves the generalization of the distilled images. 
This trend extends all the way to (and likely beyond) 512$\times$512 images, allowing us to generate high-quality distilled images at higher resolutions than ever before.
Since \glad{} acts as a plug-and-play addition to any dataset distillation method, future works can use it to increase the generality of their data and generate higher-resolution images.

\vspace{5pt}\myparagraph{Limitations.}
\looseness=-1
 Introducing StyleGAN-XL to the distillation pipeline creates a massive new memory sink.
 Our checkpointing trick allows us to mitigate this issue to some extent.
 However, this also comes at the expense of a second forward pass through the generator.
 Given that a single pass through the generator is time-consuming, a second pass doubles the overhead.
 Additionally, a large enough synthetic set requires passes through the generator to be done in multiple batches, further increasing the extra time needed.
 
Fortunately, \glad{} is compatible with any differentiable generative model, so the development of more efficient generative models in the future will naturally reduce the cost of our method as well.
 
\myparagraph{Acknowledgments.}
This work was partly done by George Cazenavette during his study at CMU. 
We thank David Charatan, Ana Dodik, and Vincent Sitzmann of MIT's Scene Representation Group for feedback on the early drafts of this work. This work is supported, in part, by the NSF Graduate Research Fellowship and J.P. Morgan Chase Faculty Research Award.

\newpage

{\small
\bibliographystyle{ieee_fullname}
\bibliography{11_references}
}

\ifarxiv \clearpage \appendix
\label{sec:appendix}
\everypar{\looseness=-1}
\section{More Visualizations}
\looseness=-1
Please see our web page for more visualizations: \href{https://georgecazenavette.github.io/glad/}{\color{blue}georgecazenavette.github.io/glad}.
\everypar{\looseness=-1}

\section{StyleGAN Latent Spaces}\lblsec{latent_explain}
Here we further elaborate on our embedding spaces for those unfamiliar with the StyleGAN architecture. 
Some details will be left out for easier digestion. 
Visualizations of these embedding spaces can be seen in \reffig{style_spaces}.

\looseness=-1
For standard image generation with StyleGAN 3 \mbox{\cite{sg3}} (and other StyleGAN models), a random vector is first sampled from the multi-variate standard normal distribution: $z \sim \mathcal{N}(0,I)$. 
This random Gaussian vector is then fed through a ``mapping'' network (typically a simple MLP) to obtain a ``style code'' W. 
In a class-conditional StyleGAN, this ``mapping'' network is the \mbox{\textit{only}} place where the class information is used. 
This W is then passed to every ``style block'' of the ``synthesis'' network as used to modulate the convolutional layers of each block. 
The ``synthesis'' network takes in a learned constant as input and uses the modulated convolutions of the style blocks to generate the realistic image.

For each distilled sample in W\textsuperscript{+} space, we optimize a \textit{different} W code for each style block and use the synthesis network to generate our synthetic data. 
The ``mapping'' network is not used aside from initializing W.
In F$n$ space, we \textit{directly} optimize the feed-forward input to the $n$\textsuperscript{th} style block as well as the W codes for all subsequent style blocks. 
Any preceding style blocks and W codes are simply ignored.
\begin{figure*}
    \centering
    \includegraphics[width=0.20\linewidth]{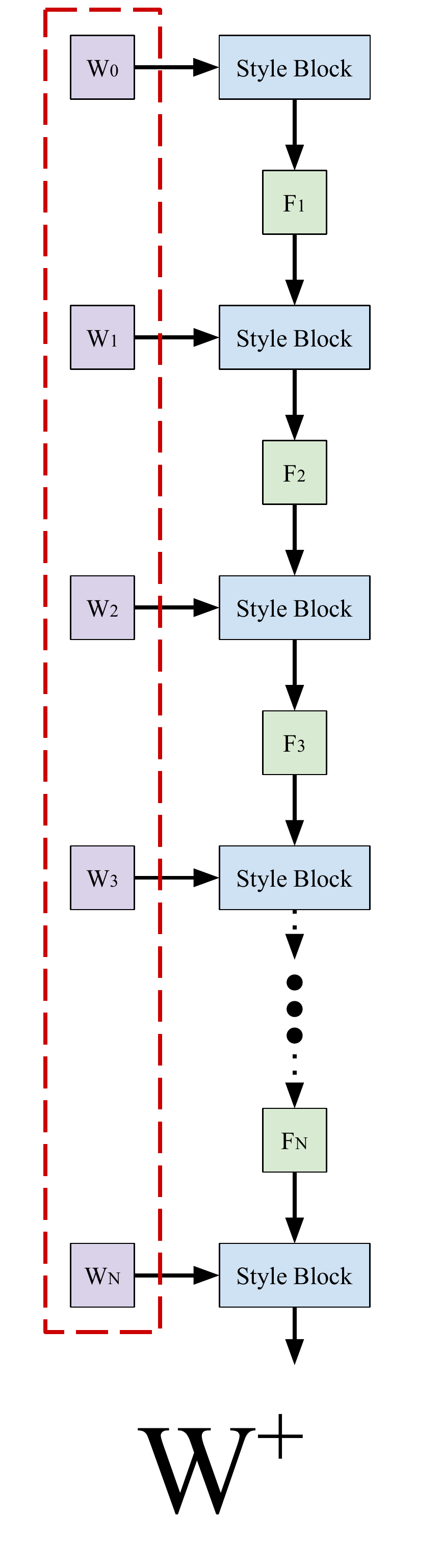}\includegraphics[width=0.20\linewidth]{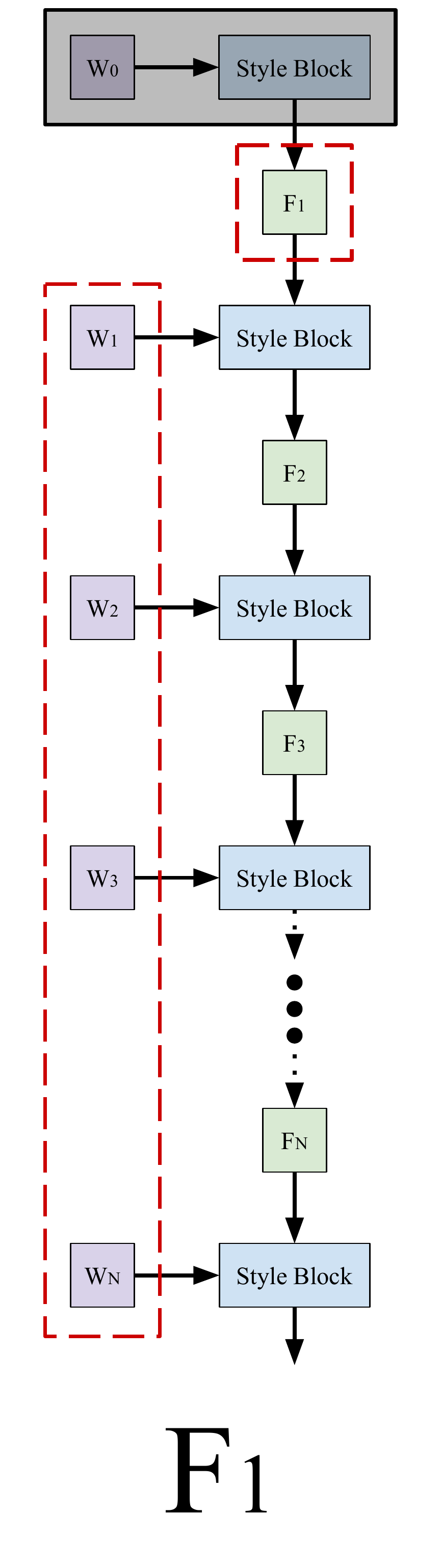}\includegraphics[width=0.20\linewidth]{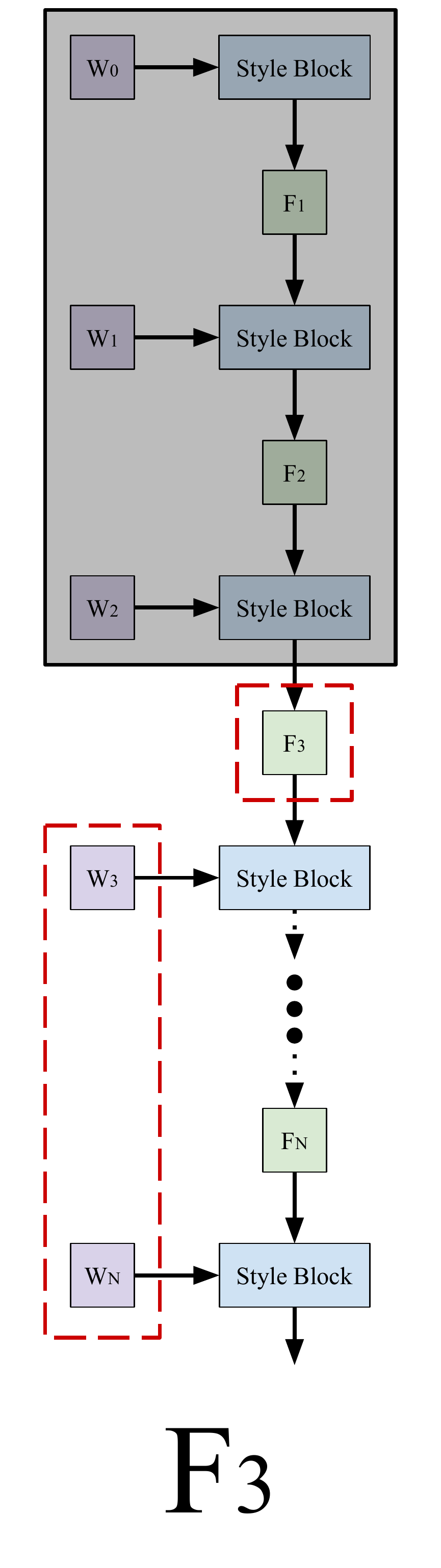}
    \caption{Different optimization spaces of StyleGAN. Latent variables boxed in red are directly optimized while those that are grayed out are not used at all. Note: the ``mapping'' network is omitted here since we do not use it in any of our optimization spaces.}
    \label{fig:style_spaces}
\end{figure*}
\vspace{-0.5cm}
\section{Dataset Specifications}
\begin{table*}[]
\scriptsize
\setlength{\tabcolsep}{3pt}
\begin{tabular}{c|cccccccccc}
Dataset    & 0                & 1                & 2               & 3                  & 4              & 5           & 6               & 7           & 8                  & 9                \\ \hline
ImageNet-A      & Leonberg          & \makecell{Probiscis\\Monkey}         & Rapeseed           & \makecell{Three-Toed\\Sloth}            & Cliff Dwelling   & \makecell{Yellow\\Lady's Slipper}  & Hamster          & Gondola     & Orca         & Limpkin         \\ \hline
ImageNet-B     & Spoonbill        & Website           & Lorikeet      & Hyena             & Earthstar          & Trollybus & Echidna             & Pomeranian & Odometer           & \makecell{Ruddy\\Turnstone}     \\ \hline
ImageNet-C     & Freight Car      & Hummingbird        & Fireboat      & Disk Brake             & Bee Eater       & Rock Beauty   & Lion        & \makecell{European\\Gallinule}         & Cabbage Butterfly            & Goldfinch \\\hline
ImageNet-D       & Ostrich     & Samoyed           & Snowbird         & \makecell{Brabancon\\Griffon}              & Chickadee    & Sorrel   & Admiral       & \makecell{Great\\Gray Owl}   & Hornbill              & Ringlet           \\\hline
ImageNet-E  & Spindle          & Toucan        & Black Swan      & \makecell{King\\Penguin} & Potter's Wheel         & Photocopier         & Screw & Tarantula  & Sscilloscope & Lycaenid            \\\hline
ImageNette    & Tench            & \makecell{English\\Springer} & \makecell{Cassette\\Player} & Chainsaw           & Church         & French Horn & \makecell{Garbage\\Truck}   & Gas Pump    & Golf Ball          & Parachute          \\\hline
ImageWoof       & \makecell{Australian \\ Terrier}        & Border Terrier       & Samoyed     & Beagle        & Shih-Tzu   & \makecell{English\\Foxhound}        & \makecell{Rhodesian\\Ridgeback}           & Dingo     & Golden Retriever       & \makecell{English\\Sheepdog}             \\\hline
ImageNet-Birds      & Peacock          & Flamingo              & Macaw        & Pelican             & \makecell{King\\Penguin}     & Bald Eagle     & Toucan      & Ostrich    & Black Swan              & Cockatoo           \\\hline
ImageNet-Fruits     & Pineapple         & Banana       & Strawberry         & Orange          & Lemon & Pomegranate  & Fig     & Bell Pepper & Cucumber        & \makecell{Granny Smith\\Apple}      \\\hline
ImageNet-Cats & \makecell{Tabby\\Cat}        & \makecell{Bengal\\Cat}       & \makecell{Persian\\Cat}     & \makecell{Siamese Cat}        & \makecell{Egyptian\\Cat}   & Lion        & Tiger           & Jaguar     & \makecell{Snow\\Leopard}       & Lynx
\end{tabular}
\caption{Class listings for our ImageNet subsets. Visualizations show classes in the same order given here.}
\vspace{0.2cm}
\label{tab:classes}
\end{table*}
Our high-resolution data is taken directly from the ImageNet 1k dataset \cite{deng2009imagenet} using PyTorch's built-in ImageNet loader \cite{pytorch}. To train our expert trajectories, we use data from the ImageNet training set. To compile \textit{our} training set for the expert trajectories, we select the classes from the given subset, resize the short side of the image to the given resolution, and take a center crop according to the given resolution (as done by \texttt{MTT} \citep{mtt}). The validation set is obtained in the same way as the ImageNet validation set.

For an enumeration of which ImageNet classes are in each of our datasets, please see Table \ref{tab:classes}.
\section{More Experimental Results}
In Table \ref{tab:convnet}, we show the performance of the distilled images on the backbone architecture (the architecture used for distillation). We note that we do not expect \glad{} to perform better than the baseline pixel-based distillation since \glad{} is designed to \textit{reduce} overfitting to the distilling architecture. 
Despite this, \dc{} and \dm{} with \glad{} perform as well or better than the pixel-based versions (with \mtt{} performing somewhat worse).

In Table \ref{tab:10ipc}, we show results using 10 distilled images per class. \glad{} still tends to perform better than pixel space distillation.

In Table \ref{tab:whole}, we show the baseline results of training each architecture on the whole dataset. We did not spend time tuning the hyper-parameters here, perhaps explaining why ConvNet tends to have the best performance.
\begin{table*}[]
\centering
\resizebox{\textwidth}{!}{%
\begin{tabular}{lccccccccccc}
Distil. Alg.                                              & Distil. Space & ImNet-A            & ImNet-B            & ImNet-C            & ImNet-D            & ImNet-E            & ImNette            & ImWoof    & ImNet-Birds & ImNet-Fruits       & ImNet-Cats         \\\midrule
                                                          & Pixel         & 51.7{\color{gray}$\pm$0.2} & 53.3{\color{gray}$\pm$1.0 }& 48.0{\color{gray}$\pm$0.7 }& 43.0{\color{gray}$\pm$0.6} & 39.5{\color{gray}$\pm$0.9 }& 41.8{\color{gray}$\pm$0.6 }& 22.6{\color{gray}$\pm$0.6} & 37.3{\color{gray}$\pm$0.8 }  & 22.4{\color{gray}$\pm$1.1 }   & 26.6{\color{gray}$\pm$0.4 } \\
\multirow{-2}{*}{\mtt{} \cite{mtt}}                       & \texttt{GLaD} (Ours)    & 50.7{\color{gray}$\pm$0.4}& 51.9{\color{gray}$\pm$1.3 }& 44.9{\color{gray}$\pm$0.4} & 39.9{\color{gray}$\pm$1.7 }& 37.6{\color{gray}$\pm$0.7} & 38.7{\color{gray}$\pm$1.6 }& 23.4{\color{gray}$\pm$1.1} & 35.8{\color{gray}$\pm$1.4 }  & 23.1{\color{gray}$\pm$0.4  }  & 26.0{\color{gray}$\pm$1.1 } \\ \midrule
\cellcolor[HTML]{FFFFFF}                                  & Pixel         & 43.2{\color{gray}$\pm$0.6 }& 47.2{\color{gray}$\pm$0.7} & 41.3{\color{gray}$\pm$0.7 }& 34.3{\color{gray}$\pm$1.5} & 34.9{\color{gray}$\pm$1.5} & 34.2{\color{gray}$\pm$1.7} & 22.5{\color{gray}$\pm$1.0} & 32.0{\color{gray}$\pm$1.5}   & 21.0{\color{gray}$\pm$0.9 }   & 22.0{\color{gray}$\pm$0.6}  \\
\multirow{-2}{*}{\cellcolor[HTML]{FFFFFF}\dc{} \cite{dc}} & \texttt{GLaD} (Ours)    & 44.1{\color{gray}$\pm$2.4 }& 49.2{\color{gray}$\pm$1.1} & 42.0{\color{gray}$\pm$0.6} & 35.6{\color{gray}$\pm$0.9} & 35.8{\color{gray}$\pm$0.9} & 35.4{\color{gray}$\pm$1.2} & 22.3{\color{gray}$\pm$1.1 }& 33.8{\color{gray}$\pm$0.9 }  & 20.7{\color{gray}$\pm$1.1 }   & 22.6{\color{gray}$\pm$0.8 } \\ \midrule
\cellcolor[HTML]{FFFFFF}                                  & Pixel         & 39.4{\color{gray}$\pm$1.8 }& 40.9{\color{gray}$\pm$1.7} & 39.0{\color{gray}$\pm$1.3} & 30.8{\color{gray}$\pm$0.9} & 27.0{\color{gray}$\pm$0.8} & 30.4{\color{gray}$\pm$2.7} & 20.7{\color{gray}$\pm$1.0} & 26.6{\color{gray}$\pm$2.6 }  & 20.4{\color{gray}$\pm$1.9  }  & 20.1{\color{gray}$\pm$1.2}  \\
\multirow{-2}{*}{\cellcolor[HTML]{FFFFFF}\dm{} \cite{dm}} & \texttt{GLaD} (Ours)   & 41.0{\color{gray}$\pm$1.5} & 42.9{\color{gray}$\pm$1.9} & 39.4{\color{gray}$\pm$0.7 }& 33.2{\color{gray}$\pm$1.4} & 30.3{\color{gray}$\pm$1.3} & 32.2{\color{gray}$\pm$1.7} & 21.2{\color{gray}$\pm$1.5} & 27.6{\color{gray}$\pm$1.9 }  & 21.8{\color{gray}$\pm$1.8  }  & 22.3{\color{gray}$\pm$1.6 }
\end{tabular}
}
\caption{
Performance on ConvNet (architecture used to distill).
}
\label{tab:convnet}
\end{table*}
\begin{table}[]
\centering
\resizebox{\linewidth}{!}{%

\begin{tabular}{lcccccc}
Distil. Alg.                                    & Distil. Space  & ImNet-A                    & ImNet-B                    & ImNet-C                    & ImNet-D                    & ImNet-E                    \\ \midrule
                                                & Pixel          & 52.3{\color{gray}$\pm$0.7} & 45.1{\color{gray}$\pm$8.3} & 40.1{\color{gray}$\pm$7.6} & 36.1{\color{gray}$\pm$0.4} & 38.1{\color{gray}$\pm$0.4} \\ 
\multirow{-2}{*}{\cellcolor[HTML]{FFFFFF}\dc{}} & \glad{} (Ours) & \textbf{53.1{\color{gray}$\pm$1.4}} & \textbf{50.1{\color{gray}$\pm$0.6}} & \textbf{48.9{\color{gray}$\pm$1.1}} & \textbf{38.9{\color{gray}$\pm$1.0}} & \textbf{38.4{\color{gray}$\pm$0.7}} \\ \midrule
\cellcolor[HTML]{FFFFFF}                        & Pixel          & 52.6{\color{gray}$\pm$0.4} & 50.6{\color{gray}$\pm$0.5} & 47.5{\color{gray}$\pm$0.7} & 35.4{\color{gray}$\pm$0.4} & 36.0{\color{gray}$\pm$0.5} \\
\multirow{-2}{*}{\cellcolor[HTML]{FFFFFF}\dm{}} & \glad{} (Ours) & \textbf{52.8{\color{gray}$\pm$1.0}} & \textbf{51.3{\color{gray}$\pm$0.6}} & \textbf{49.7{\color{gray}$\pm$0.4}} & \textbf{36.4{\color{gray}$\pm$0.4}} & \textbf{38.6{\color{gray}$\pm$0.7}}
\end{tabular}
}

\caption{Performance with 10 images/class.
}
\label{tab:10ipc}
\end{table}
\begin{table}[]

\centering
\resizebox{\linewidth}{!}{%
\begin{tabular}{lccccc}
         Arch.
         & ImNet-A                                            & ImNet-B                                            & ImNet-C                                            & ImNet-D                                            & ImNet-E                                            \\\midrule
ConvNet  & \cellcolor[HTML]{FFFFFF}90.6{\color{gray}$\pm$0.6} & \cellcolor[HTML]{FFFFFF}92.3{\color{gray}$\pm$0.2} & \cellcolor[HTML]{FFFFFF}84.2{\color{gray}$\pm$0.3} & \cellcolor[HTML]{FFFFFF}74.5{\color{gray}$\pm$1.0} & \cellcolor[HTML]{FFFFFF}76.2{\color{gray}$\pm$0.6} \\
ResNet18 & \cellcolor[HTML]{FFFFFF}78.8{\color{gray}$\pm$1.6} & \cellcolor[HTML]{FFFFFF}80.2{\color{gray}$\pm$1.1} & \cellcolor[HTML]{FFFFFF}69.2{\color{gray}$\pm$1.6} & \cellcolor[HTML]{FFFFFF}51.0{\color{gray}$\pm$0.7} & \cellcolor[HTML]{FFFFFF}53.2{\color{gray}$\pm$2.8} \\
VGG11    & \cellcolor[HTML]{FFFFFF}78.4{\color{gray}$\pm$1.1} & \cellcolor[HTML]{FFFFFF}81.4{\color{gray}$\pm$1.5} & \cellcolor[HTML]{FFFFFF}74.6{\color{gray}$\pm$1.2} & \cellcolor[HTML]{FFFFFF}67.3{\color{gray}$\pm$1.6} & \cellcolor[HTML]{FFFFFF}67.8{\color{gray}$\pm$1.3} \\
AlexNet  & \cellcolor[HTML]{FFFFFF}81.0{\color{gray}$\pm$0.3} & \cellcolor[HTML]{FFFFFF}76.5{\color{gray}$\pm$1.4} & \cellcolor[HTML]{FFFFFF}72.2{\color{gray}$\pm$1.1} & \cellcolor[HTML]{FFFFFF}65.4{\color{gray}$\pm$1.1} & \cellcolor[HTML]{FFFFFF}63.5{\color{gray}$\pm$1.1} \\
ViT      & \cellcolor[HTML]{FFFFFF}77.5{\color{gray}$\pm$0.4} & \cellcolor[HTML]{FFFFFF}76.4{\color{gray}$\pm$0.4} & \cellcolor[HTML]{FFFFFF}75.5{\color{gray}$\pm$1.4} & \cellcolor[HTML]{FFFFFF}58.6{\color{gray}$\pm$0.9} & \cellcolor[HTML]{FFFFFF}59.5{\color{gray}$\pm$1.2}
\end{tabular}
}

\caption{Training networks from scratch on the \textit{whole} dataset.
}
\label{tab:whole}
\end{table}
\section{Hyper-Parameters and Experimental Details}
For the experiments on \texttt{MTT} and our new method, we base our experiments on the open-source code for
 \dc{}+\dm{} (\href{https://github.com/VICO-UoE/DatasetCondensation}{link}) \cite{dc,dm}, 
 \mtt{} (\href{https://github.com/GeorgeCazenavette/mtt-distillation}{link}) \cite{mtt}, and
 \tesla{} (\href{https://openreview.net/forum?id=dN70O8pmW8}{link}) \cite{tesla}.

To optimize the distilled images/latents and learnable synthetic step size ($\alpha$), we use the same optimizer and hyper-parameters as the original methods. For the W\textsuperscript{+} latents, divide the learning rate by 10.

For our \texttt{MTT} experiments, we set the number of synthetic steps per iteration ($N$) as 10, the number of real epochs to match ($M$) as 2, and the maximum starting epoch ($T^+$) set to 2. 
All experiments on $\texttt{MTT}$ and our new method are run for 5k iterations and then evaluated via the protocol described in the body of the paper.

All $32\times32$, $128\times128$, $256\times 256$, and $512\times512$ experiments are distilled using ConvNetD3, ConvNetD5, ConvNetD6, and ConvNetD7 respectively as the backbone.

The same suite of differentiable augmentations (originally from the \texttt{DSA} codebase \cite{dsa}) is used for all experiments: color, crop, cutout, flip, scale, and rotate with the default parameters.

To obtain the expert trajectories used by \texttt{MTT}, we train a model from scratch on the real dataset for 15 epochs of SGD with a learning rate of $10^{-2}$, a batch size of 256, and NO momentum or regularization.

Our experiments were run on a combination of RTX2080ti, RTX3090, RTX6000, RTXA5000, and RTXA6000 GPUs depending on availability.
\clearpage
 \fi

\end{document}